\documentclass[journal,onecolumn]{IEEEtran}
\pdfoutput = 1
\usepackage{times}
\usepackage{epsfig}
\usepackage{epstopdf}
\usepackage{graphicx}
\usepackage{amsmath}
\usepackage{amssymb}
\usepackage{subcaption}
\usepackage{color}
\usepackage{algpseudocode}
\usepackage{algorithm}
\usepackage{placeins}
\usepackage{rotating}
\usepackage{capt-of,etoolbox}
\newcommand{\argmin}{\operatornamewithlimits{argmin}}
\DeclareMathOperator{\tr}{tr}
\DeclareMathOperator{\prox}{prox}
\DeclareMathOperator{\diag}{diag}
\DeclareMathOperator{\sign}{sgn}

\newcommand{\vassilis}[1]{{\textcolor[rgb]{1,0,0}{#1}}}

\usepackage[pagebackref=true,breaklinks=true,letterpaper=true,colorlinks,bookmarks=false]{hyperref}
\ifCLASSINFOpdf

\else
 
\fi

\begin{document}

\title{Robust Principal Component Analysis on Graphs}
\author{Nauman Shahid$^{*}$, Vassilis Kalofolias$^{*}$, Xavier Bresson$^{*}$, Michael Bronstein$^{\dagger}$, Pierre Vandergheynst$^{*}$\\
$^{*}$Signal Processing Laboratory (LTS2), EPFL, Switzerland, Email: firstname.lastname@epfl.ch \\
$^\dagger$ Universita della Svizzera Italiana, Switzerland, Email: michael.bronstein@usi.ch
}


\maketitle

\begin{abstract}
 
    Principal Component Analysis (PCA) is the most widely used tool for linear dimensionality reduction and clustering. Still it is highly sensitive to outliers and does not scale well with respect to the number of data samples. Robust PCA solves the first issue with a sparse penalty term. The second issue can be handled with the matrix factorization model, which is however non-convex. Besides, PCA based clustering can also be enhanced by using a graph of data similarity. In this article, we introduce a new model called `Robust PCA on Graphs' which incorporates spectral graph regularization into the Robust PCA framework. Our proposed model benefits from 1) the robustness of principal components to occlusions and missing values, 2) enhanced low-rank recovery, 3) improved clustering property due to the graph smoothness assumption on the low-rank matrix, and 4) convexity of the resulting optimization problem. Extensive experiments on 8 benchmark, 3 video and 2 artificial datasets with  corruptions clearly reveal that our model outperforms 10 other state-of-the-art models in its clustering and low-rank recovery tasks.
\end{abstract}

\begin{figure*}[htbp]
    \centering
        \centering
        \includegraphics[width=1.0\textwidth]{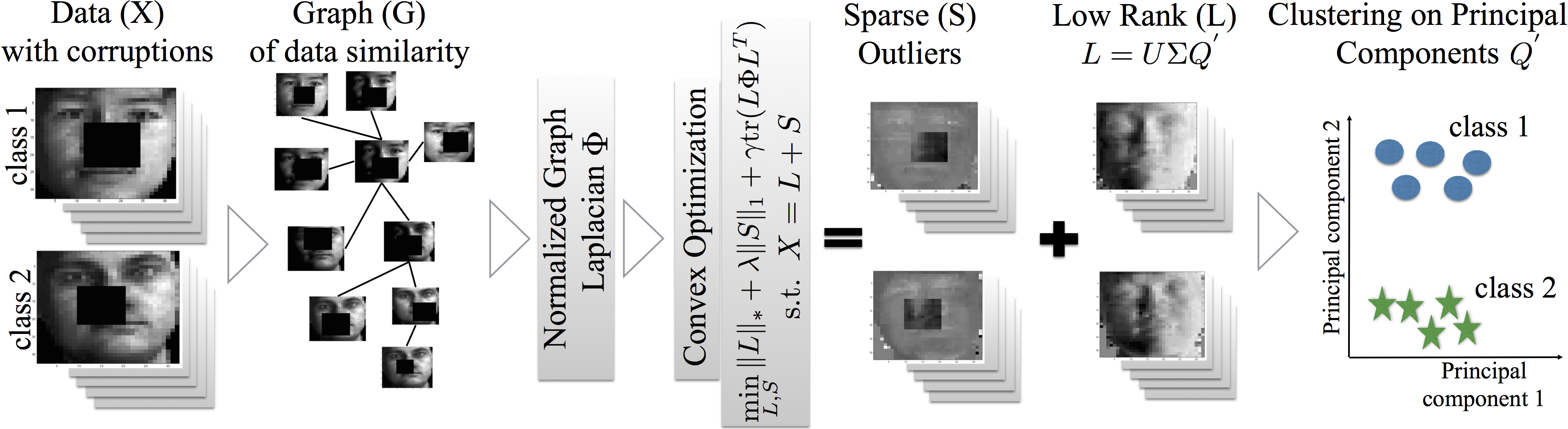}
         \caption{Main idea of our work. }
        \label{fig:main_idea}
    \end{figure*}

\section{Introduction}\label{sec:intro}

\textbf{What is PCA?} Given a data matrix $X \in \mathbb{R}^{p\times n}$ with $n$ p-dimensional data vectors, the classical PCA can be formulated as learning the projection $Q \in \mathbb{R}^{d\times n}$ of $X$ in a $d$-dimensional linear space characterized by an orthonormal basis $U \in \mathbb{R}^{p\times d}$ ($1^{st}$ model in Table~\ref{tab:models}). Traditionally, $U$ and $Q$ are termed as \textit{principal directions} and \textit{principal components} and the product $UQ $ is known as the \textit{low-rank approximation $L \in \mathbb{R}^{p\times n}$} of $X$. 


\textbf{Applications of PCA}: PCA has been widely used for two important applications.
\begin{enumerate}

\item Dimensionality reduction or {low-rank} recovery. 

\item Data clustering using the  {principal components} $Q$ in the low dimensional space.
\end{enumerate}

However, classical PCA formulation is susceptible to errors in the data because of the quadratic term. Thus, an outlier in the data might result in erratic {principal components $Q$} which can effect both the dimensionality reduction and clustering.

\textbf{Robust dimensionality reduction using PCA}: Candes et. al. \cite{candes2011robust} demonstrated that PCA can be made robust to outliers by exactly recovering the {low-rank representation $L$} even from grossly corrupted data $X$ by solving a simple convex problem, named Robust PCA (RPCA, $2^{nd}$ model in Table~\ref{tab:models}). In this model, the data corruptions are represented by the sparse matrix $S \in \mathbb{R}^{p\times n}$. Extensions of this model for inexact recovery \cite{5513535} and outlier pursuit \cite{xu2010robust}  have been proposed.

\textbf{The clustering quality of PCA can be significantly improved} by incorporating the data manifold information in the form of a graph $G$ \cite{jiang2013graph,zhang2013low,jin2014low,tao2014low}.  The underlying assumption is that the low-dimensional embedding of data $X$ lies on  a smooth manifold \cite{belkin2003laplacian}. Let $G = (V,E,A)$ be the graph  with vertex set $V$ as the samples of $X$, $E$ the set of pairwise edges between the vertices $V$ and $A$ the adjacency matrix which encodes the weights of the edges $E$. Then, the \textit{normalized graph Laplacian} matrix $\Phi \in \mathbb{R}^{n\times n}$ which characterizes the graph $G$ is defined as $\Phi = D^{-1/2}(D - A)D^{-1/2} = I - D^{-1/2}AD^{-1/2}$, where $D$ is the degree matrix defined as $D = \diag(d_{i})$ and $d_{i}=\sum_{j}A_{ij}$. Assuming that a p-nearest neighbors graph is available, several methods to construct $A$ have been proposed in the literature.  The three major weighting methods are: 1) binary, 2) heat kernel  and 3) correlation distance \cite{jin2014low}. 

\textbf{In this paper} we propose a novel convex method to improve the {clustering and low-rank recovery} applications of PCA by incorporating spectral graph regularization to the Robust PCA framework.  Extensive experiments reveal that  our proposed model is more robust to occlusions and missing values as compared to 10 state-of-the-art dimensionality reduction models.

  \begin{table*}[htbp]
\footnotesize
\caption{A comparison of various PCA models and their properties. $X \in \mathbb{R}^{p\times n}$ is the data matrix, $U \in \mathbb{R}^{p\times d}$ and $Q \in \mathbb{R}^{d\times n}$ are the {principal directions} and  {principal components} in a $d$ dimensional linear space (rank = $d$). $L = UQ \in \mathbb{R}^{p\times n}$ is the  {low-rank representation} and $S \in \mathbb{R}^{p\times n}$ is the sparse matrix. $\Phi$, $\Phi^{g}$ and $\Phi_{h}^{g} \in \mathbb{S}^{n\times n}$ characterize a simple graph or a hypergraph $G$ between the samples of $X$. $\|\cdot\|_{F}$, $\|\cdot\|_{*}$ and $\|\cdot\|_{1}$ denote the Frobenius, nuclear and $l_{1}$ norms respectively.}
\centering
\resizebox{1.0\textwidth}{!}{\begin{tabular}[t]{| c | c | c | c | c | c | c | c |} \hline
   &   \textbf{Model}   & \textbf{Objective}  &   \textbf{Constraints}   & \textbf{Parameters}   &  \textbf{Graph?} & \textbf{Factors?}  & \textbf{Convex?}   \\\hline
1 &  PCA    &  $ \min_{U,Q} \|X-UQ\|_{F}^{2}$  &   $U^{T}U  = I$ & $d$ & no  &    yes &  no   \\\hline
2   & RPCA \cite{candes2011robust}  &  $\min_{L,S} \|L\|_{*} + \lambda\|S\|_{1}$  &   $X = L + S$  & $\lambda$  & no  &  no & yes  \\\hline
\vassilis{\textbf{3}}   & \vassilis{\textbf{PROPOSED}}  & \vassilis{$ \mathbf{\min_{L,S} \|L\|_{*} + \lambda\|S\|_{1} + \gamma \tr (L\Phi L^{T})} $}  & \vassilis{$\mathbf{X = L + S}$} & \vassilis{$\mathbf{\lambda, \gamma}$}    &  \vassilis{\textbf{YES}}&  \vassilis{\textbf{NO}}   & \vassilis{\textbf{YES}} \\\hline
4  &  GLPCA  \cite{jiang2013graph}   & $\min_{U,Q} \|X-UQ\|_{F}^{2}  + \gamma \tr(Q\Phi Q^{T})$ & $QQ^{T} = I $  &     &  &     & \\\cline{1-3}
5  & RGLPCA  \cite{jiang2013graph} & $\min_{U,Q} \|X-UQ\|_{2,1}  + \gamma \tr(Q\Phi Q^{T})$ &  &  $d,\gamma$  & &   &  \\\cline{1-4}
6  & MMF  \cite{zhang2013low}      &  $\min_{U,Q} \|X-UQ\|_{F}^{2}  + \gamma \tr(Q\Phi Q^{T})$ & $U^{T}U = I $   &   & yes & yes & no \\\cline{1-5}
7  & MMMF   \cite{tao2014low} & $\min_{U,Q,\boldsymbol{\alpha}} \|X-UQ\|_{F}^{2}  + \gamma\tr(Q(\sum_{g}{\alpha}_{g}\Phi^{g}) Q^{T}) + \beta\|\boldsymbol{\alpha}\|^{2}$ & $U^{T}U = I $ & $d,\gamma,\beta$ &  &   &  \\\cline{1-3}
8  & MHMF  \cite{jin2014low}  & $\min_{U,Q,\boldsymbol{\alpha}} \|X-UQ\|_{F}^{2}  + \gamma\tr(Q(\sum_{g}{\alpha}_{g}\Phi_{h}^{g}) Q^{T}) + \beta\|\boldsymbol{\alpha}\|^{2}$ &  $\boldsymbol{1}^{T}\boldsymbol{\alpha} = \boldsymbol{1}$  &   &  &  & \\\hline
\end{tabular}}
\label{tab:models}
\end{table*}

\section{Main Idea \& Proposed Model}\label{sec:mainidea}
The main contributions of our work are:

\begin{enumerate}
\item Exact recovery of the {low-rank representation $L$} from grossly corrupted  data $X$.

\item Recovery of the {low-rank representation $L$} that also reveals the underlying class separation.

\item High cluster purity in the low dimensional space even when no clean data is available. 

\item  A simple convex optimization problem with minimal parameters to achieve these objectives.
\end{enumerate}

The figure on page 1 illustrates the  main idea of our work.  

Without any doubt the contributions 1 and 4 are given by the work of Candes et. al. \cite{candes2011robust}, namely Robust PCA (RPCA, $2^{nd}$ model in Table~\ref{tab:models}). Thus, as a first step, we propose that instead of utilizing a classical PCA-like model so as to explicitly learn {principal directions $U$} and {principal components $Q$} of data $X$, one can directly recover the {low-rank matrix $L$} itself ($L = UQ$). 

Secondly and more importantly, to achieve contributions 2 and 3 we propose: \textit{The low-rank matrix $L$ itself lies on a smooth manifold and it can be recovered directly on this manifold}.  \textbf{Our proposed model} is as follows:
\begin{align}\label{eq:RobPCA}
& \min_{L,S} \|L\|_{*} + \lambda\|S\|_{1} + \gamma \tr (L\Phi L^{T}), \\\nonumber
&  \text{s.t.} ~ X = L + S,
\end{align} 
where the sparse errors in the data are modeled by $S$ and $L$ is the {low-rank} approximation of $X$. Parameters $\lambda$ and $\gamma$ control the amount of sparsity of $S$ and smoothness of $L$ on the graph $\Phi$ respectively. We will define our graph in Section~\ref{sec:Lap}.

\textbf{Generalization of Robust PCA}:
We call our proposed model \eqref{eq:RobPCA} {Robust PCA on Graphs}. It is a direct extension of the Robust PCA proposed by Candes et. al. \cite{candes2011robust} with smoothness manifold regularization. That is, setting $\gamma = 0$ in our model~\eqref{eq:RobPCA} we obtain the standard model of \cite{candes2011robust}.


 \section{Related Works: Factorized PCA Models}\label{sec:relatedwork} 

Both manifold regularization and robustness techniques for PCA have been proposed in the literature, either separately or combined \cite{jiang2013graph,zhang2013low,jin2014low,tao2014low}. Following the classical PCA model they explicitly learn two factors $U$ and $Q$ such that $X \approx UQ$. We will, therefore refer to these models as \textit{factorized PCA models}. Furthermore, unlike our model, some of these works \cite{jiang2013graph} assume {the graph smoothness of principal components $Q$ (instead of $L = UQ$)}. 

Jiang et. al. proposed Graph Laplacian PCA (GLPCA) \cite{jiang2013graph} ($4^{th}$ model in Table~\ref{tab:models}) which leverages the graph regularization of {principal components $Q$} using the term $\tr(Q\Phi Q^{T})$ for clustering in the low dimensional space. They also proposed a robust version of their model ($5^{th}$ model in Table~\ref{tab:models}) and demonstrated the robustness of {principal components $Q$} to occlusions in the data.
 
 Zhang and Zhao \cite{zhang2013low} proposed Manifold Regularized Matrix Factorization (MMF, $6^{th}$ model in Table~\ref{tab:models}) which exploits the orthonormality constraint on the {principal directions $U$} (contrary to \cite{jiang2013graph}) to acquire a unique {low-rank matrix $L = UQ$} for any optimal pair $U$, $Q$. In this case we have  $\tr(Q\Phi Q^{T}) = \tr(UQ\Phi(UQ)^{T}) = \tr(L\Phi L^{T})$, therefore this model implicitly assumes the graph smoothness of $L$.  The extensions of this model with an ensemble of graph and hypergraph regularization terms have been proposed by Tao et. al. \cite{tao2014low} and Jin et. al. \cite{jin2014low} respectively ($7^{th}$ and $8^{th}$ models in Table~\ref{tab:models}). 


\textbf{Shortcomings of state-of-the-art}:
Although the models proposed in \cite{jiang2013graph,zhang2013low,jin2014low,tao2014low} leverage the graph to learn enhanced class structures, they still suffer from numerous problems. Most of these models  are not robust to data corruptions \cite{zhang2013low,jin2014low,tao2014low}. Those which leverage the robustness  suffer from non-convexity  \cite{jiang2013graph}. An ensemble of graphs or hypergraphs leverages the non-linearity of data in an effective manner \cite{jin2014low,tao2014low}. However, it makes the models  non-convex and the resulting alternating direction methods can get stuck in local minima. 

 \textbf{Notation \& Terminology}: Throughout this article $\|\cdot\|_{F}$, $\|\cdot\|_{*}$ and $\|\cdot\|_{1}$ denote the Frobenius, nuclear and $l_{1}$ norms respectively. We will refer to the regularization term $\tr(Q\Phi Q^{T})$ as \textit{principal components graph regularization}. Note that the graph regularization involves {principal components $Q$}, not {principal directions $U$}. We will also refer to the regularization term $\tr(L\Phi L^{T})$ as \textit{low-rank graph regularization}. RPCA and our proposed models ($2^{nd}$ and $3^{rd}$ models in Table~\ref{tab:models}) which perform exact low-rank recovery by splitting $X = L + S$ will be referred to as \textit{non-factorized PCA models}. A comparison of various PCA models introduced so far is presented in Table~\ref{tab:models}. Note that only RPCA and our proposed model leverage convexity and enjoy a unique global optimum with guaranteed convergence. 

\section{Comparison with Related Works}
The main differences between our model~\eqref{eq:RobPCA} and the various state-of-the-art factorized PCA models \cite{jiang2013graph,zhang2013low,lu2013graph,jin2014low,tao2014low} are, as summarized in Table~\ref{tab:models}, the following.

 \textbf{Non-factorized model}: Instead of explicitly learning the {principal directions $U$} and principal components $Q$, it learns their product, i.e. the {low-rank matrix $L$}.  Hence,~\eqref{eq:RobPCA} is a {non-factorized PCA model}.

 \textbf{Exact low-rank recovery}: Unlike factorized models we target the {exact low-rank} recovery by modeling the data matrix as the sum of {low-rank $L$} and a {sparse matrix $S$}.

\textbf{Different graph regularization term}:  Model~\eqref{eq:RobPCA} is based on the assumption that it is the {low-rank matrix} $L$ that is smooth on the graph, and not just the {principal components} matrix $Q$. Therefore we replace the {principal components graph} term $\tr(Q\Phi Q^\top)$ with the {low-rank graph} term $\tr(L\Phi L^\top)$. Note that as explained in Section~\ref{sec:relatedwork}, the two terms are only equivalent if orthogonality of $U$ is further assumed, as in \cite{zhang2013low,lu2013graph,jin2014low,tao2014low} and not in \cite{jiang2013graph}.

\subsection{Advantages over Factorized PCA Models}

 \textbf{Robustness to gross corruptions for clustering \& low-rank recovery}: The {low-rank graph $\tr (L\Phi L^{T})$} can be more realistic than the {principal components graph $\tr (Q\Phi Q^{T})$}. It allows the {principal directions $U$} to benefit from the graph regularization as well (recall that $L = UQ$). Thus, our model enjoys an enhanced {low-rank} recovery  and class separation even from grossly corrupted data. For details, please refer to Sections~\ref{sec:clustering} \&~\ref{sec:recovery} (also see Fig.~\ref{fig:basisvectors} in supplementary material). 

 \textbf{Convexity}: It is a strictly convex problem and a unique global optimum can be obtained by using standard methods like an Alternating Direction Method of Multipliers (ADMM) \cite{boyd2010distributed}.

\textbf{One model parameter only}: Our model does not require the rank of $L$ to be specified up-front. The nuclear norm
relaxation enables the automatic selection of an appropriate rank based on the parameters
$\lambda$ and $\gamma$. Furthermore, as illustrated in our experiments, the value
$\lambda = 1/\sqrt{\max(n,p)}$ proposed in \cite{candes2011robust}  gives very good results. As a result,
the only unknown parameter to be selected is $\gamma$, and for this we can use methods
such as cross validation (For additional details please see Figs.~\ref{fig:images_ce} \&~\ref{fig:images_lowe} in the supplementary material).

 \section{Optimization Solution}\label{sec:opt}
We use an ADMM \cite{boyd2010distributed} to rewrite Problem \eqref{eq:RobPCA}:
\begin{align}\label{eq:RobPCA}
& \min_{L,S,W} \|L\|_{*} + \lambda\|S\|_{1} + \gamma \tr (W\Phi W^{T}) \nonumber \\
&  \text{s.t.} ~ X = L + S , ~ L = W. \nonumber 
\end{align}
Thus, the augmented Lagrangian and iterative scheme are:
\begin{align*}
(L,S,W)^{k+1} & =  \argmin_{L,S,W} \|L\|_{*} + \lambda\|S\|_{1}  + \gamma \tr (W\Phi W^{T}) \nonumber \\
& + \langle Z^{k}_{1}, X - L - S\rangle \ + \frac{r_{1}}{2}\|X-L-S\|_{F}^{2}\nonumber\\
&  + \langle Z^{k}_{2},W-L\rangle+\frac{r_{2}}{2}\|W-L\|_{F}^{2},  \nonumber \\
Z_{1}^{k+1} & =  Z_{1}^{k} + r_{1}(X-L^{k+1}-S^{k+1}), \nonumber \\
 Z_{2}^{k+1} & =  Z_{2}^{k} + r_{2}(W^{k+1}-L^{k+1}), \nonumber 
\end{align*}
where $Z_{1} \in \mathbb{R}^{p\times n}$ and $Z_{2} \in \mathbb{R}^{p\times n}$ are the lagrange multipliers and $k$ is the iteration index. Let $ H^{k}_{1} = X-S^{k}+Z^{k}_{1}/r_{1}$ and $ H^{k}_{2} = W^{k} + Z^{k}_{2}/r_{2}$, then this reduces to the following updates for $L$, $S$ and $W$ as:
\begin{align*}
& {L^{k+1}} = \prox_{\frac{1}{(r_{1}+r_{2})} \| L\|_{*}}\Big(\frac{r_{1}H^{k}_{1}+r_{2}H^{k}_{2}}{r_{1}+r_{2}}\Big), \\
& S^{k+1} = \prox_{{\frac{\lambda}{r_{1}}}\|S\|_{1}}\Big(X-L^{k+1}+\frac{Z^{k}_{1}}{r_{1}}\Big), \\
& W^{k+1} = r_{2}(\gamma \Phi + r_{2}I)^{-1}\Big(L^{k+1} - \frac{Z^{k}_{2}}{r_{2}}\Big), \\
\end{align*}
where $\prox_{f}$ is the proximity operator  of the convex function $f$ as defined in \cite{combettes2011proximal}. The details of this solution, algorithm, convergence and  computational complexity are given in the supplementary material Sections~\ref{sec:admm_sol},~\ref{sec:admm_algo} \&~\ref{sec:convergence}.

\section{Experimental Setup}\label{sec:setup}
We use the model~\eqref{eq:RobPCA} to solve two major PCA-based problems.

\begin{enumerate}
\item {Data clustering in low-dimensional space} with corrupted and uncorrupted data (\textbf{Section~\ref{sec:clustering} }). 

\item {Low-rank recovery} from corrupted data (\textbf{Section~\ref{sec:recovery}}). 
\end{enumerate}

Our extensive experimental setup is designed to test the robustness and generalization capability of our model to a wide variety of datasets and corruptions for the above two applications. {{Precisely, we perform our experiments on 8 benchmark, 3 video and 2 artificial datasets with  10 different types of corruptions and compare with 10 state-of-the-art dimensionality  reduction models}} as explained in sections~\ref{sec:clustering_setup} \&~\ref{sec:lowrank_setup}.

\subsection{Setup for Clustering}\label{sec:clustering_setup}

\subsubsection{Datasets} 

All the datasets are well-known benchmarks. \textbf{The 6 image databases} include CMU PIE\footnotemark[1]\footnotetext[1]{\href{http://vasc.ri.cmu.edu/idb/html/face/}{http://vasc.ri.cmu.edu/idb/html/face/}}, ORL\footnotemark[2]\footnotetext[2]{\href{http://www.cl.cam.ac.uk/research/dtg/attarchive/facedatabase.html}{http://www.cl.cam.ac.uk/research/dtg/attarchive/facedatabase.html}}, YALE\footnotemark[3]\footnotetext[3]{\href{http://vision.ucsd.edu/content/yale-face-database}{http://vision.ucsd.edu/content/yale-face-database}}, COIL20\footnotemark[4]\footnotetext[4]{\href{http://www.cs.columbia.edu/CAVE/software/softlib/coil-20.php}{http://www.cs.columbia.edu/CAVE/software/softlib/coil-20.php}}, MNIST\footnotemark[5]\footnotetext[5]{\href{http://yann.lecun.com/exdb/mnist/}{http://yann.lecun.com/exdb/mnist/}}   and USPS data sets. \textbf{MFeat database} \footnotemark[6]\footnotetext[6]{\href{https://archive.ics.uci.edu/ml/datasets/Multiple+Features}{https://archive.ics.uci.edu/ml/datasets/Multiple+Features}}  consists of features extracted from handwritten numerals and the \textbf{BCI database}\footnotemark[7]\footnotetext[7]{\href{http://olivier.chapelle.cc/ssl-book/benchmarks.html}{http://olivier.chapelle.cc/ssl-book/benchmarks.html}} comprises of features extracted from a Brain Computer Interface setup. Our choice of datasets is  based on their various properties such as pose changes, rotation (for digits), data type and  non-negativity, as presented in Table~\ref{tab:datasets} in the supplementary material.

\subsubsection{Comparison with 10 models} 

We compare the clustering performance of our model with k-means on the original data $X$ and {{9 other dimensionality reduction models}}. These models  can be divided into two categories. 

\textbf{1. Models without graph}: 1) classical Principal Component Analysis (PCA)  2) Non-negative Matrix Factorization (NMF) \cite{lee1999learning}  and 4) Robust PCA (RPCA) \cite{candes2011robust}.

 \textbf{2. Models with graph}: These models can be further divided into two categories based on the graph type. \textbf{a. Principal components graph}: 1)  Normalized Cuts (NCuts) \cite{shi2000normalized}, 2) Laplacian Eigenmaps (LE) \cite{belkin2003laplacian},  3) Graph Laplacian PCA (GLPCA) \cite{jiang2013graph}, 4) Robust Graph Laplacian PCA (RGLPCA) \cite{jiang2013graph},  5) Manifold Regularized Matrix Factorization (MMF) \cite{zhang2013low},  6) Graph Regularized Non-negative Matrix Factorization (GNMF) \cite{cai2011graph}, \textbf{b. Low-rank graph}: Our proposed model. 

Table~\ref{tab:datasets} and Fig.~\ref{fig:venn}  in the supplementary material give a summary of all the models.


\subsubsection{Corruptions in datasets} 

\textbf{Corruptions in image databases}: We  introduce three types of corruptions in each of the 6 image databases:

 \textbf{1. No corruptions}.

\textbf{2. Fully corrupted data}. Two types of corruptions are introduced in {all} the images of each database: a) {Block occlusions} ranging from 10 to 40\% of the image size. b) {Missing pixels} ranging from 10\% to 40\% of the total pixels in the image. These corruptions are modeled by placing zeros uniformly randomly in the images.

\textbf{3. Sample specific corruptions}. The above two types of corruptions (occlusions and missing pixels) are introduced in only 25\% of the images of each database.


\textbf{Corruptions in non-image databases}: We introduce only {full and sample specific missing values}  in the non-image databases because the block occlusions in non-image databases correspond to an unrealistic assumption. Example missing pixels and block occlusions in the image are shown in Fig.~\ref{fig:corruptions} in the supplementary material.

\textbf{Pre-processing}: For NCuts, LE, PCA, GLPCA, RGLPCA, MMF, RPCA and our model we pre-process the datasets to zero mean and unit standard deviation along the features. Additionally for MMF all the samples are made unit norm as suggested in \cite{zhang2013low}. For NMF and GNMF we only pre-process the non-negative datasets to unit norm along the samples. We perform pre-processing after introducing the corruptions.

\subsubsection{Clustering Metric}

We use \textit{{clustering error}} to evaluate the clustering performance of all the models considered in this work. The {clustering error} is $E = (\frac{1}{n}\sum_{r=1}^{k}n_{r})\times 100 $, where $n_{r}$ is the number of misclassified samples in cluster $r$. We report the minimum clustering error from 10 runs of k-means (k = number of classes) on the {principal components $Q$}.  This procedure  reduces the bias introduced by the non-convex nature of k-means. For RPCA and our model we obtain {principal components} $Q^{'}$ via SVD of the low-rank matrix $L = U\Sigma Q^{'}$ during the nuclear proximal update in every iteration.  {For more details of the clustering error evaluation and parameter selection scheme for each of  the models, please refer to Fig.~\ref{fig:flow_clustering} and Table~\ref{tab:models_param} of the supplementary material}.

\subsection{Setup for Low-Rank Recovery}\label{sec:lowrank_setup}

 Since the {low-rank} ground truth for the 8 benchmark datasets used for clustering is not available, we perform the following two types of experiments.

\begin{enumerate}
\item \textbf{Quantitative evaluation} of the normalized {low-rank} reconstruction error using corrupted artificial datasets.

\item \textbf{Visualization} of the recovered {low-rank representations} for 1) occluded images of the CMU PIE dataset and 2) static background of 3 different videos\footnotemark[8]\footnotetext[8]{\href{https://sites.google.com/site/backgroundsubtraction/test-sequences}{https://sites.google.com/site/backgroundsubtraction/test-sequences}}.
\end{enumerate} 

\subsection{Normalized Graph Laplacian}\label{sec:Lap}

In order to construct the graph Laplacian $\Phi$, the pairwise Euclidean distance is computed between each pair of the vectorized data samples $(x_{i},x_{j})$. Let $\Omega$ be the matrix which contains all the pairwise distances, then $\Omega_{ij}$, the Euclidean distance between $x_{i}$ and $x_{j}$ is given as:

\textbf{Case I: Block Occlusions}
\begin{equation*}
\Omega_{ij} = \sqrt{\frac{\|M_{ij} \circ(x_{i} - x_{j})\|^{2}_{2}}{\|M_{ij}\|_{1}}},
\end{equation*} 
where $M_{ij} \in \{0,1\}^{p}$ is the vector mask corresponding to the intersection of  uncorrupted values in $x_{i}$ and $x_{j}$. Thus
\begin{equation*}
M^{l}_{ij} =
\left\{
\begin{array}{l l}
     1 & \text{if features} \hspace{0.2cm}  x^{l}_{i}  \hspace{0.2cm} \& \hspace{0.2cm}   x^{l}_{j} \hspace{0.2cm} \text{observed},\\
      0 & \text{otherwise}
   \end{array}
   \right.
\end{equation*}
Thus, we detect the block occlusions and consider only the observed pixels. 

\textbf{Case II: Random Missing Values}: We use a Total Variation denoising procedure and calculate $\Omega_{ij}$ using the cleaned images.

Let $\omega_{min}$ be the minimum of all the pairwise distances in $\Omega$. Then the adjacency matrix $A$ for the graph $G$ is constructed by using  
\begin{equation*}
A_{ij} = \exp\Big(-\frac{(\Omega_{ij}-\omega_{min})^{2}}{\sigma^{2}}\Big)
\end{equation*}

Finally, the normalized graph Laplacian $\Phi = I - D^{-1/2}AD^{-1/2}$ is calculated, where $D$ is the degree matrix.  Note that different types of data might call for different distance metrics and values of $\sigma^{2}$, however for all the experiments and datasets used in this work the Euclidean distance metric and $\sigma^{2} = 0.05$ work well. To clarify further for the reader, we present a detailed example  of graph construction  with corrupted and uncorrupted images in Fig.~\ref{fig:laplacians} of supplementary material.

\section{Clustering in Low Dimensional Space}\label{sec:clustering}

\begin{figure*}[htbp]
    \centering
        \centering
        \includegraphics[width=1.0\textwidth]{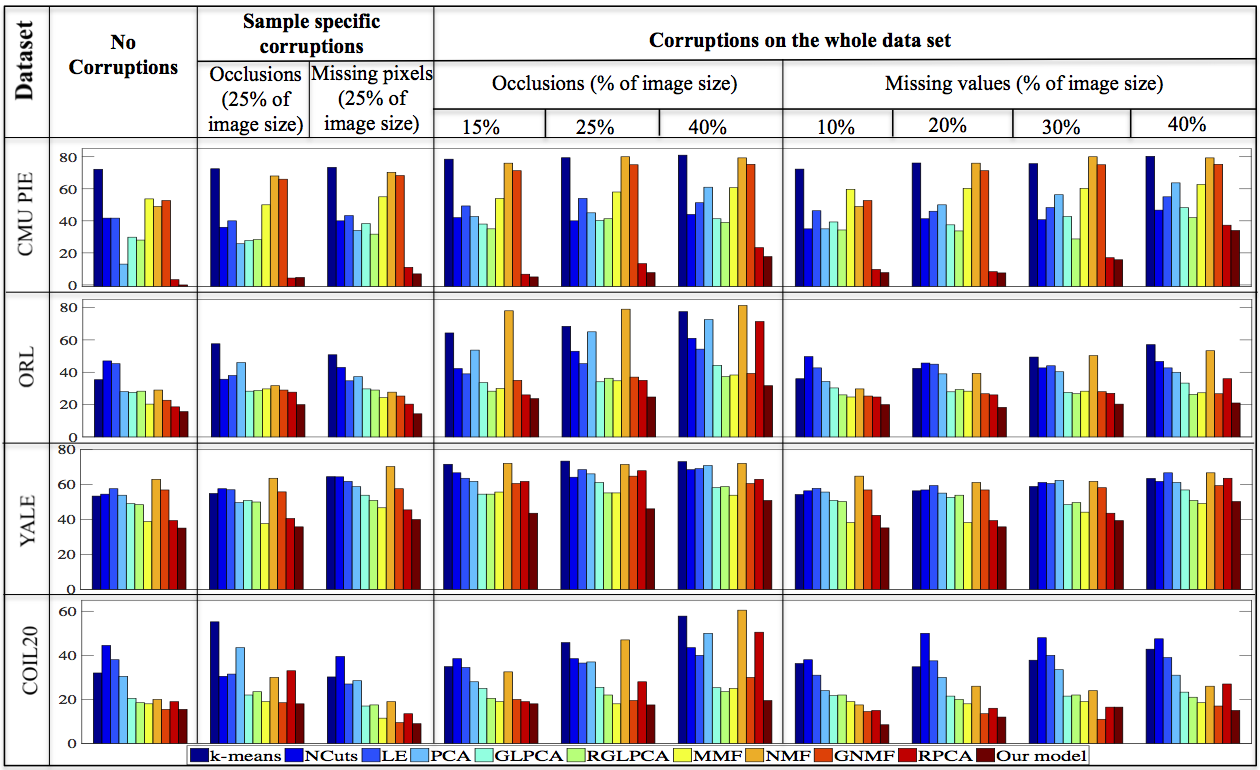}
         \caption{Clustering error of  various dimensionality reduction models for CMU PIE, ORL, YALE \& COIL20 datasets. The compared models are: 1) k-means without dimensionality reduction 2) Normalized Cuts (NCuts) \cite{shi2000normalized} 3) Laplacian Eigenmaps (LE) \cite{belkin2003laplacian}  4) classical Principal Component Analysis (PCA)  5) Graph Laplacian PCA (GLPCA) \cite{jiang2013graph} 6) Robust Graph Laplacian PCA (RGLPCA) \cite{jiang2013graph}  7) Manifold Regularized Matrix Factorization (MMF) \cite{zhang2013low}  8) Non-negative Matrix Factorization \cite{lee1999learning}  9) Graph Regularized Non-negative Matrix Factorization (GNMF) \cite{cai2011graph}, 10) Robust PCA (RPCA) \cite{candes2011robust} and 11) Robust PCA on Graphs (proposed model). Two types of full and partial corruptions are introduced in the data: 1) Block occlusions and 2) Random missing values. The numerical errors corresponding to this figure along with additional results on MNIST, USPS, MFeat and BCI datasets are presented in Tables~\ref{tab:results1},~\ref{tab:results1a} \&~\ref{tab:results2} of the supplementary material.}
        \label{fig:images_error}
    \end{figure*}

\begin{figure*}[htbp]
    \centering
        \centering
        \includegraphics[width=1.0\textwidth]{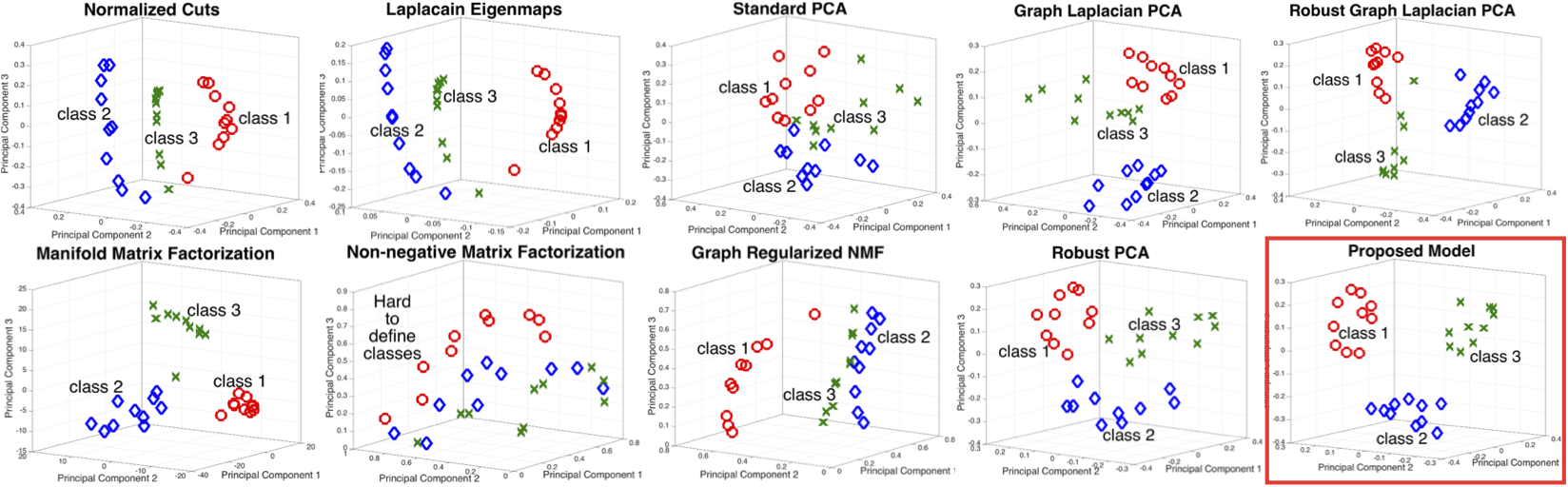}
         \caption{Principal components $Q$ of three classes of CMU PIE data set in 3 dimensional space. Only ten instances of each class are used with block occlusions occupying 20\% of the entire image. Our proposed model (on the lower right corner) gives well-separated {principal components} without any clustering error.}
        \label{fig:images_PCS}
    \end{figure*}

Fig.~\ref{fig:images_error}  presents experimental results for the first important application of our model, i.e. clustering.  Fig.~\ref{fig:images_PCS} illustrates the {principal components $Q$} for three classes of occluded CMU PIE data set in 3-dimensional space. Our model outperforms others in most of the cases with different types and levels of occlusions (please refer to Tables~\ref{tab:results1},~\ref{tab:results1a} \&~\ref{tab:results2} in the supplementary material for additional results). In the next few paragraphs we elaborate further on 1) the advantage of graph over non-graph models, 2) the advantage of {low-rank graph} over {principal components graph}. Throughout the description of our results, the comparison with RPCA is of specific interest because it is a special case of our proposed model.

\subsection{Is the Graph Useful?}\label{sec:graph_useful} 
 Our model performs better than k-means, standard PCA, NMF and RPCA which do not leverage the graph.
 
\textbf{Example case I: CMU PIE database with no pose variation}: Consider  the case of CMU PIE dataset in Fig.~\ref{fig:images_error}. This dataset does not suffer from pose changes and we observe that our model attains as low as $0\%$ error when there are no corruptions in the data. Furthermore, we attain lowest error even with the increasing levels of occlusions and missing pixels. This can also be observed visually from Fig.~\ref{fig:images_PCS} where the principal components for our model are better separated than others.

\textbf{Example case II: COIL20 database with pose variation}: Our model outperforms  RPCA and other non-graph models also for the COIL20 database which suffers from significant pose changes. Thus, even a graph constructed using the simple scheme of Section~\ref{sec:Lap}  enhances the robustness of our model to gross data corruptions and pose changes. Similar conclusions can be drawn for all the other databases as well.

\subsection{Low-Rank  {or} Principal Components Graph?}\label{sec:which_graph}

We compare the performance of our model with NCuts, LE, GLPCA, RGLPCA, MMF \& GNMF which use {principal components graph}. It is obvious from Fig.~\ref{fig:images_error}  that our model outperforms the others even for the datasets with pose changes. Similar conclusions can be drawn for all the other databases with corruptions and by visually comparing the {principal components} of these models in Fig.~\ref{fig:images_PCS} as well. 

Unlike factorized models, the {principal directions $U$} in the {low-rank graph $\tr(L\Phi L^{T})$}  benefit from the graph regularization as well and show robustness to corruptions.  This leads to a better clustering in the low-dimensional space even when the graph is constructed from corrupted data (please refer to  Fig.~\ref{fig:basisvectors} of the supplementary material for further experimental explanation).

\subsection{Robustness to Graph Quality}

\begin{figure}[htbp]
    \centering
        \centering
        \includegraphics[width=0.45\textwidth]{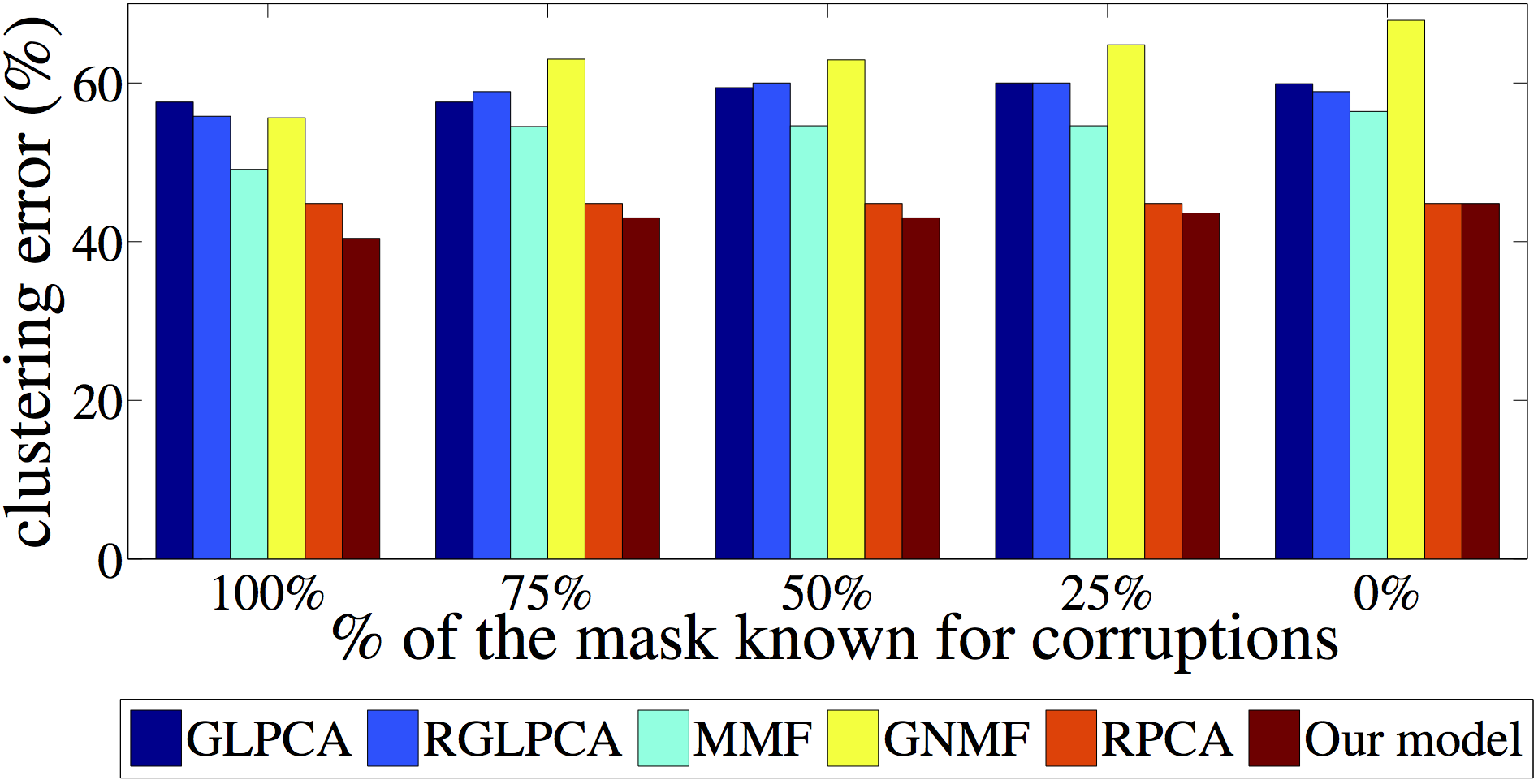}
         \caption{Effect of lack of information about the corruptions mask on the clustering error. YALE dataset  with 35\% random missing pixels is used for this experiment. Our model (last bar in each group) outperforms the others.} 
        \label{fig:image_badgraph}
    \end{figure}

We perform an experiment on the YALE dataset with 35\% random missing pixels. Fig.~\ref{fig:image_badgraph} shows the variation in clustering error of different PCA models by using a graph constructed with decreasing information about the mask. Our model still outperforms others even though the quality of graph deteriorates with corrupted data. It is essential to note that in the worst case scenario when the mask is not known at all, our model performs equal to RPCA but still better than those which use a {principal components graph}.

%

\section{Low-Rank Recovery}\label{sec:recovery}

\subsection{Low-Rank Recovery from Artificial Datasets}\label{sec:recovery_artificialdata}

To perform a quantitative comparison of {exact low-rank recovery} between RPCA and our model, we perform experiments on 2 artificial datasets as in \cite{candes2011robust}. Note that only RPCA and our model perform {exact low-rank recovery} so we do not present results for other PCA models. We generate low-rank square matrices $L = A^{T}B \in \mathbb{R}^{n \times n}$ (rank $d$ = $0.02n$ to $0.3n$) with $A$ and $B$ independently chosen $d\times n$ matrices with i.i.d. Gaussian entries of zero mean and variance $1/n$. We  introduce 6\% to 30\% errors $\rho = \|S\|_{0}/n^{2}$ in these matrices from an i.i.d. Bernoulli distribution for support of $S$ with two sign schemes.
\begin{enumerate}

\item {Random signs}: Each corrupted entry takes a value $\pm 1$ with a probability $\rho/2$.

\item {Coherent signs}: The sign for each corrupted entry is coherent with the sign of the corresponding entry in $L$.
\end{enumerate}

\begin{figure*}[htbp]
    \centering
     \begin{subfigure}[b]{0.23\textwidth}
        \centering
        \includegraphics[width=1.0\textwidth]{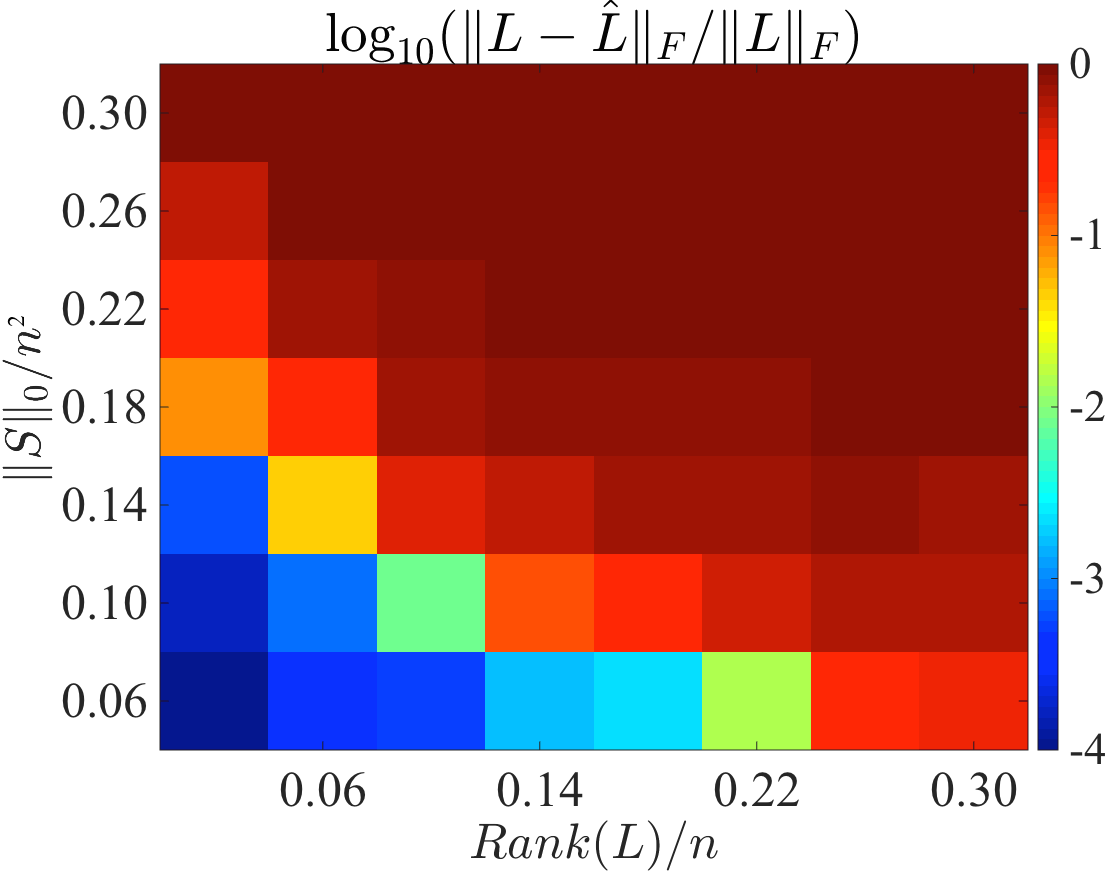}
        \caption{Our model, random signs}
    \end{subfigure}
    \hfill
    \begin{subfigure}[b]{0.23\textwidth}
        \centering
        \includegraphics[width=1.0\textwidth]{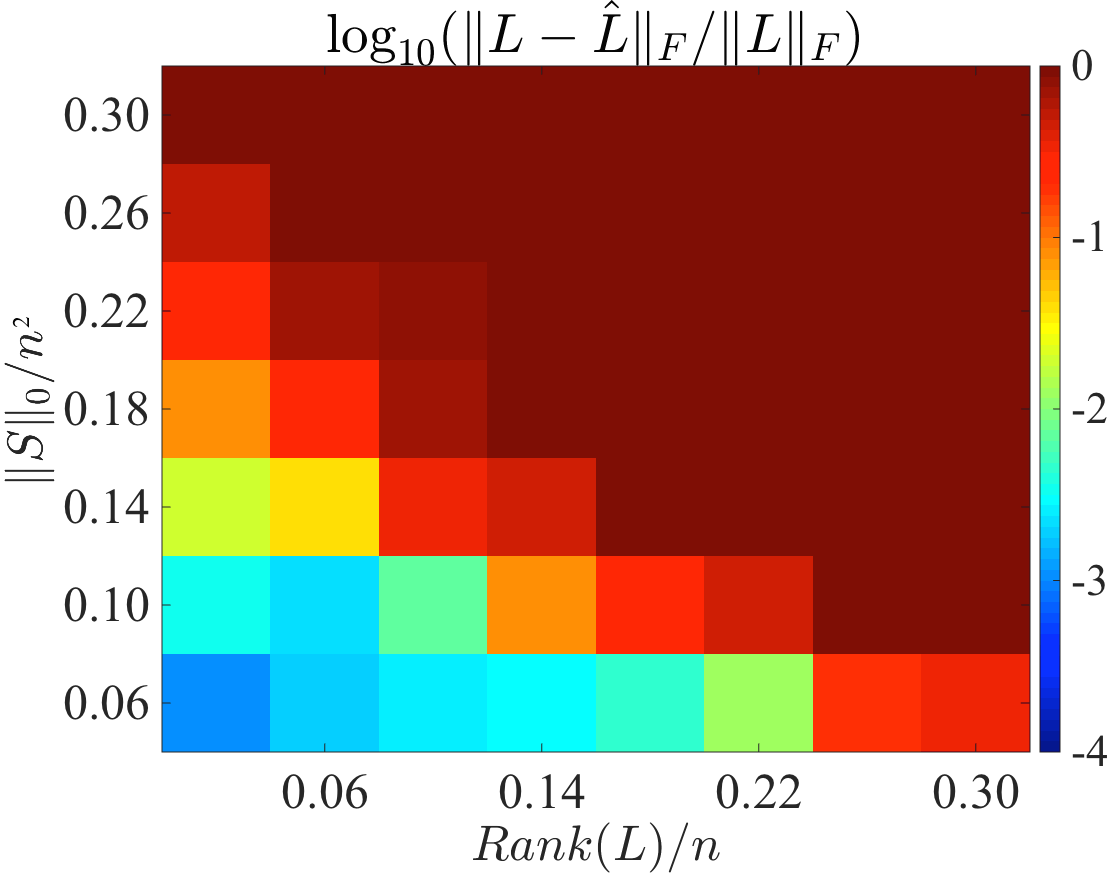}
        \caption{RPCA, random signs}
    \end{subfigure}
    \hfill
    \begin{subfigure}[b]{0.23\textwidth}
        \centering
        \includegraphics[width=1.0\textwidth]{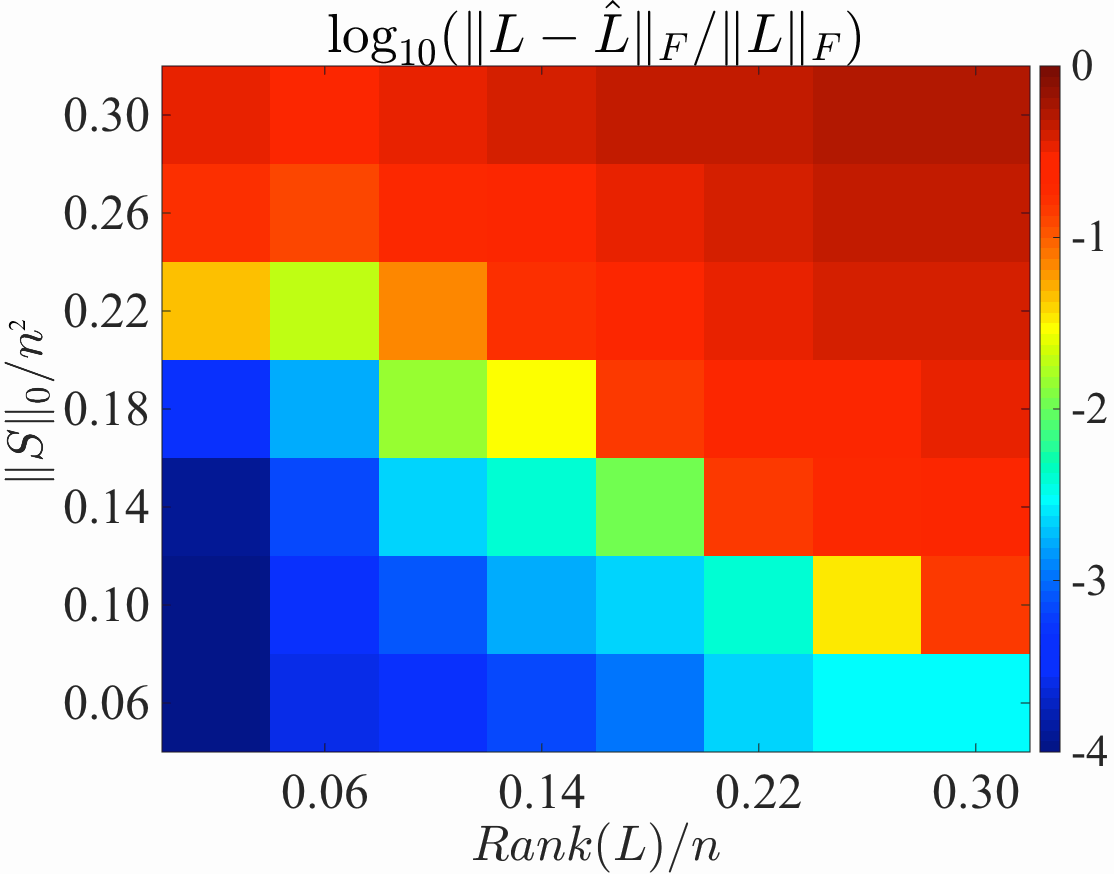}
        \caption{Our model, coherent signs}
    \end{subfigure}
    \hfill
    \begin{subfigure}[b]{0.23\textwidth}
        \centering
        \includegraphics[width=1.0\textwidth]{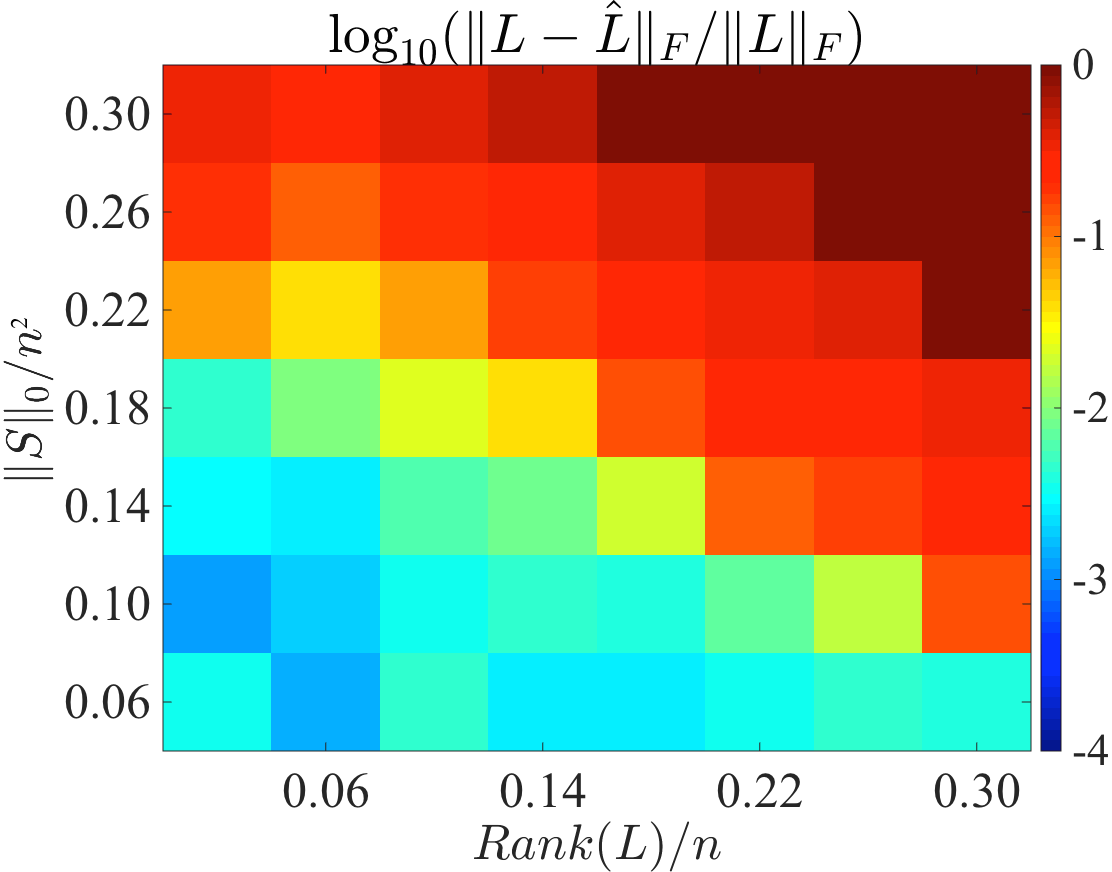}
       \caption{RPCA, coherent signs}
    \end{subfigure}
    \caption{Variation of Log normalized low-rank reconstruction error on artificial data (Section~\ref{sec:recovery_artificialdata}) with (rank, error). The larger and darker the blue area, the lower reconstruction error of the model.}
     \label{fig:errorartificial}
\end{figure*}

Fig.~\ref{fig:errorartificial} compares the variation of log normalized low-rank reconstruction error $\log(\|L-\hat{L}\|_{F}/\|L\|_{F})$ with rank (x-axis) and error (y-axis) between RPCA (b and d) and our model (a and c). The larger and darker the blue region, the better is the reconstruction. The first two plots correspond to the {random sign} scheme and the next ones to {coherent sign} scheme.  It can be seen that for each (rank,error) pair the reconstruction error for our model is less than that for RPCA. Hence, the {low-rank graph} helps in an enhanced low-rank recovery.

\subsection{Low-Rank Recovery from Corrupted Faces}\label{sec:recovery_faces}

\begin{figure*}[htbp]
    \centering
        \centering
        \includegraphics[width=1.0\textwidth]{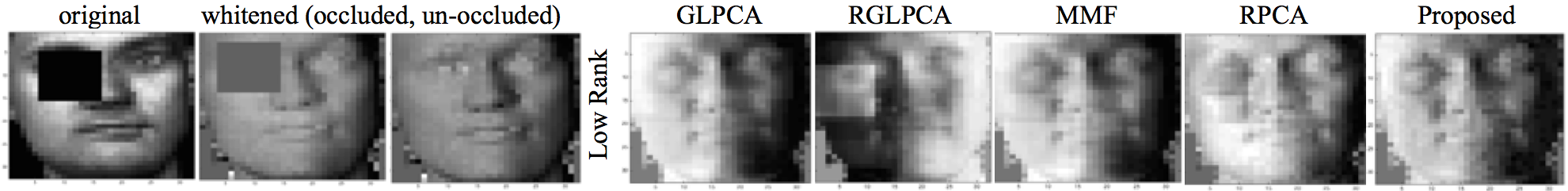}
         \caption{Clean low-rank recovery of one image of the CMU PIE data set corresponding to each of the PCA models. The images of one person are corrupted by 10\% block occlusions.  $1^{st}$ figure shows the actual occluded image, $2^{nd}$ and $3^{rd}$ figures show the whitened occluded and un-occluded images. Since PCA requires whitening, the recovered low-rank images in figures 4 to 8 should resemble the un-occluded whitened image. }
        \label{fig:images_LS}
    \end{figure*}

 We use the PCA models to recover the clean {low-rank representations} from a set of occluded images of CMU PIE dataset as shown in Fig.~\ref{fig:images_LS}.  None of the factorized models using {principal components graph} is able to perfectly separate the block occlusion from the actual image. This is not surprising because these models (except RGLPCA) are not robust to gross errors. Our model is able to separate the occlusion better than all the models. Even though it inherits the robustness from RPCA, we observe that the robust recovery of the {low-rank} representation $L$ is greatly enhanced by using the {low-rank graph}. Please refer to  Fig.~\ref{fig:imagesLSfull} in the supplementary material for more results.


\subsection{Low-Rank Background Extraction from Videos}\label{sec:recovery_videos}

\begin{figure*}[!]
    \centering
        \includegraphics[width=1.0\textwidth]{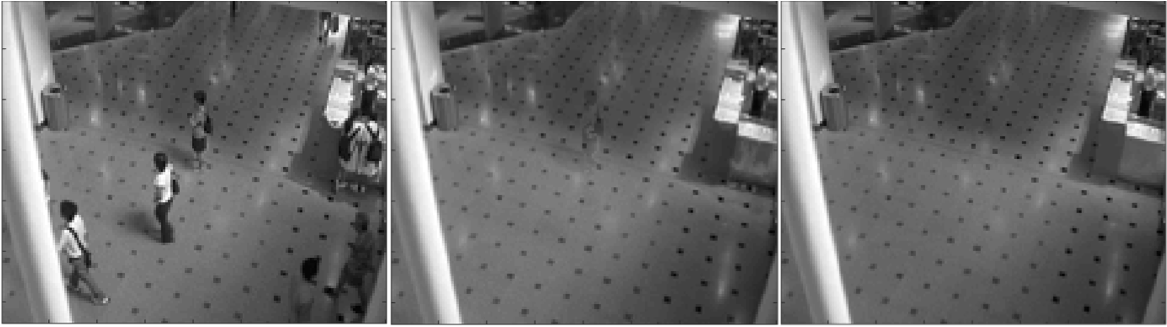}
    \caption{One frame of the recovered background from a video of a shopping mall lobby. left) actual frame, middle) recovered low-rank background using Robust PCA and right) recovered background using our proposed model. The presence of graph in our model enables better recovery and removes all the walking people from the frame. }
    \label{fig:imagesvideo}
\end{figure*}

Static background separation from the dynamic foreground is an interesting application of {low-rank recovery}. We use 3 videos \footnotemark[8] (1000 frames each) to recover the static {low-rank} background. The graph Laplacian $\Phi$ is constructed between the frames of videos without utilizing any prior information about sparse errors. The {low-rank} ground truth is not available for these   videos so we present a visual comparison between RPCA and our model for one of the frames in Fig.~\ref{fig:imagesvideo}. The RPCA model (middle) is unable to completely remove the person in the middle of the frame and his shadow from the {low-rank} frame. However, the presence of graph in our model (right) helps in a better recovery of the static background, which in this case is the empty shopping mall lobby. The pictures are best viewed on a computer screen on the electronic version of this article. \textbf{Complete videos for our model can be found here \footnotemark[9]\footnotetext[9]{\href{https://vid.me/GN0X}{https://vid.me/GN0X}} \footnotemark[10]\footnotetext[10]{\href{https://vid.me/vR6d}{https://vid.me/vR6d}} \footnotemark[11]\footnotetext[11]{\href{https://vid.me/RDgN}{https://vid.me/RDgN}}}.  For more results please refer to Fig.~\ref{fig:videosfull} of supplementary material.

\section{Parameter Selection Scheme for Our Model}\label{sec:params}

There are two parameters for our model: 1) Sparsity penalty $\lambda$ and 2) Graph regularization penalty $\gamma$. The parameter $\lambda$ can be set approximately equal to $1/\sqrt{\max(n,p)}$ where $n$ is the number of data samples and $p$ is the dimension. This simple rule, as suggested by Candes et. al. \cite{candes2011robust} works reasonably well for our {clustering} \& {low-rank recovery error}  experiments. After fixing $\lambda$, the parameter $\gamma$ can be easily selected by cross-validation.  The minimum clustering error  always occurs around this value of  $\lambda$, irrespective of the size, number of classes and $\%$ of corruptions in the datasets. Due to lack of space we present the detailed results on the variation of {clustering} \& {low-rank recovery error} over the $(\lambda,\gamma)$ grid  in Figs.~\ref{fig:images_ce} \&~\ref{fig:images_lowe} of the supplementary material.

\section{Comparison of Computation Times}

\begin{figure}[!h]
    \centering
        \centering
        \includegraphics[width=0.47\textwidth]{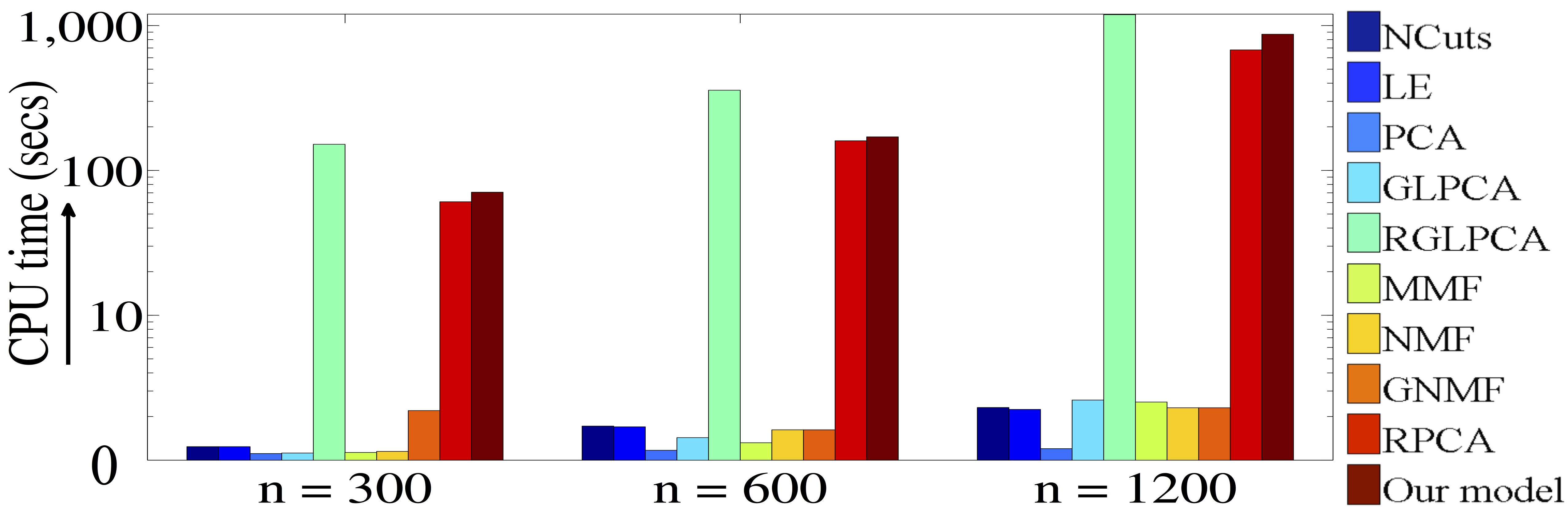}
         \caption{Computation time of different models for CMU PIE dataset with 20\% occlusions. }
        \label{fig:image_time}
    \end{figure}

We corrupt different sizes of the CMU PIE dataset (n = 300, 600 and 1200)  with 20\% occlusions and compute the time for one run of each model which gives the minimum clustering error (Fig.~\ref{fig:image_time}).  Clearly,   RGLPCA has the highest computation time, followed by our proposed model. However, the trade-off between the clustering error and computational time is worth observing from Figs.~\ref{fig:images_error} and~\ref{fig:image_time} (more details in the supplementary material Table~\ref{tab:times}). The large computation time of our model is dominated by the expensive SVD step in every iteration.  Our future work will be dedicated to reduce this cost by using randomized algorithms for SVD \cite{witten2013randomized} and exploiting the parallel processing capabilities \cite{lucas2014parallel}.

\section{Conclusion}\label{sec:conclusion}

We present `Robust PCA on Graphs', a generalization of the Robust PCA framework which leverages spectral graph regularization on the low-rank representation. The proposed model targets  exact low-rank recovery and  enhanced clustering in the low-dimensional space from grossly corrupted datasets by solving a convex optimization problem with minimal parameters. Numerical experiments on several benchmark datasets reveal that clustering using our low-rank graph regularization scheme outperforms various other state-of-the-art factorized PCA based models which use principal components graph regularization. Experiments for exact low-rank recovery from artificial datasets and static background separation from videos demonstrate its ability of to perform an enhanced low-rank recovery. On the top of that, it is also robust w.r.t. the graph construction strategy and performs well even when the graph is constructed from corrupted data.

\textbf{Note}: {Due to space constraints we present only a few important results  in the main text. However, we encourage the reader to see the supplementary material for additional results}.

{\small
\bibliographystyle{ieee}
\bibliography{pcabib.bib}
}

\onecolumn
\clearpage
\newpage
\appendix

\section{Appendices}\label{sec:A}

\subsection{Details of ADMM Solution}\label{sec:admm_sol}
We use the Alternating Direction Method of Multipliers (ADMM) \cite{boyd2010distributed}. First, the problem \eqref{eq:RobPCA} is split as follows:
\begin{align}
& \min_{L,S,W} \|L\|_{*} + \lambda\|S\|_{1} + \gamma \tr (W\Phi W^{T}) \nonumber \\
&  \text{s.t.} ~ X = L + S , ~ L = W. \nonumber 
\end{align}
Second, the augmented Lagrangian and iterative scheme are introduced:
\begin{align}\label{auglag}
(L^{k+1},S^{k+1},W^{k+1}) & = \argmin_{L,S,W} \|L\|_{*} + \lambda\|S\|_{1}  + \gamma \tr (W\Phi W^{T}) + \langle Z^{k}_{1}, X - L - S\rangle \ + \frac{r_{1}}{2}\|X-L-S\|_{F}^{2}   \nonumber \\
& + \langle Z^{k}_{2},W-L\rangle+\frac{r_{2}}{2}\|W-L\|_{F}^{2},   \\
Z_{1}^{k+1}  & = Z_{1}^{k} + r_{1}(X-L^{k+1}-S^{k+1}),\\
 Z_{2}^{k+1} & = Z_{2}^{k} + r_{2}(W^{k+1}-L^{k+1}),
\end{align}
where $Z_{1}$ and $Z_{2}$ are the lagrange multipliers (dual variables) and $k$ is the iteration index. 

In order to update the primal variables $L$, $S$ and $W$ we consider the definition of proximity operator \cite{combettes2011proximal}. Let $F \in \Gamma_{0}(\mathbb{R}^{N})$, where $\Gamma_{0}(\mathbb{R}^{N})$ is the class of lower semi-continuous convex functions from $\mathbb{R}^{N}$ to $]-\infty, +\infty]$ such that domain of F $\neq \varnothing$. Then, for every $x\in \mathbb{R}^{N}$, the following minimization problem admits a unique solution which is known as the {\textbf{proximity operator of F}} denoted by $\prox_{F}$.
\begin{equation*}
\prox_{F}(x) = \argmin_{y\in \mathbb{R}^{N}} F(y) + \frac{1}{2}\|x-y\|^{2}_{2}.
\end{equation*}

Using this definition, the updates for $L$, $S$ and $W$ at $(k+1)^{st}$ iteration using the previous iterates $L^{k},S^{k},W^{k},Z^{k}_{1},Z^{k}_{2}$ can be made using the proximity operators.

\textbf{Update $L$}

Keeping only the terms with $L$ in Eq.~\eqref{auglag} we get
\begin{align*}
L^{k+1} = & \argmin_{L} \|L\|_{*}  + \langle Z^{k}_{1}, X - L- S^{k}\rangle \ + \frac{r_{1}}{2}\|X-L-S^{k}\|_{F}^{2} + \langle Z^{k}_{2},W^{k}-L\rangle+\frac{r_{2}}{2}\|W^{k}-L\|_{F}^{2}  \nonumber \\
   = & \argmin_{L} \|L\|_{*}  + \frac{r_{1}}{2}\Big\|L-\big(X-S^{k}+\frac{Z^{k}_{1}}{r_{1}}\big)\Big\|^{2}_{F} + \frac{r_{2}}{2}\Big\|L-\big(W^{k}+\frac{Z^{k}_{2}}{r_{2}}\big)\Big\|^{2}_{F} \nonumber \\
   = & \argmin_{L} \|L\|_{*}  + \frac{r_{1}+r_{2}}{2}\Big\|L - \frac{r_{1}H^{k}_{1}+r_{2}H^{k}_{2}}{r_{1}+r_{2}}\Big\|^{2}_{F} \nonumber \\
   = & \prox_{\frac{1}{(r_{1}+r_{2})} \| L\|_{*}}\Big(\frac{r_{1}H^{k}_{1}+r_{2}H^{k}_{2}}{r_{1}+r_{2}}\Big), \nonumber
\end{align*}
where $ H^{k}_{1} = X-S^{k}+Z^{k}_{1}/r_{1}$ and $ H^{k}_{2} = W^{k} + Z^{k}_{2}/r_{2}$. Let $\Omega_{\tau} : \mathbb{R}^{N} \longrightarrow \mathbb{R}^{N}$ denote the element-wise soft-thresholding operator  $\Omega_{\tau }(x) = \sign(x)\max(|x|-\tau,0)$, then we can define $D_{\tau}(A) = P\Omega_{\tau}(\Sigma) Q^{T}$ as the singular value thresholding operator for matrix $A$, where $A = P\Sigma Q^{T}$ is any singular value decomposition of $A$. Let $A = \frac{r_{1}H^{k}_{1}+r_{2}H^{k}_{2}}{r_{1}+r_{2}}$ and $r= (r_{1}+r_{2})/2$ then 
\begin{equation}
L^{k+1} = D_{\frac{1}{r}}(A) =  P\Omega_{\frac{1}{r}}(\Sigma) Q^{T}.
\end{equation}

\textbf{Update $S$}

Following a similar procedure, we can write the update for $S$ as.

\begin{align}
 S^{k+1} & = \prox_{{\frac{\lambda}{r_{1}}}\|S\|_{1}}\Big(X-L^{k+1}+\frac{Z^{k}_{1}}{r_{1}}\Big) \nonumber \\
              & = \Omega_{\frac{\lambda}{r_{1}}}\Big(X-L^{k+1}+\frac{Z^{k}_{1}}{r_{1}}\Big) 
\end{align}

\textbf{Update $W$}

By keeping only the terms with $W$ in eq.~\eqref{auglag} we get

\begin{equation*}
W^{k+1} = \argmin_{W} \gamma \tr(W\Phi W^{T}) + \frac{r_{2}}{2}\Big\|W - \Big(L^{k+1} - \frac{Z^{k}_{2}}{2}\Big) \Big\|^{2}_{F}
\end{equation*}
which is a smooth function in $W$, so we can use the optimality condition to find a closed form solution for $W$.
\begin{equation}
 W^{k+1} = r_{2}(\gamma \Phi + r_{2}I)^{-1}\Big(L^{k+1} - \frac{Z^{k}_{2}}{r_{2}}\Big) 
 \end{equation}

Projected Conjugate Gradient method was used to update $W$.

\FloatBarrier
\subsection{Algorithm}\label{sec:admm_algo}
\begin{algorithm}
\caption{ADMM algorithm for Robust PCA on Graphs}\label{euclid}
\begin{algorithmic}[1]
\Procedure{Robust PCA Graphs}{$X \in \mathbb{R}^{p\times n},\Phi\in \mathbb{S}^{n}_{+},\lambda,\gamma$}  \Comment{inputs}
\State $k \gets 0$   \Comment{iteration index $k$}
\State $L^{k} \gets rand(n,p), \hspace{1cm} W^{k} \gets rand(n,p) , \hspace{1cm} S^{k} \gets rand(n,p)$    \Comment{Initialize primal variables}
\State $r_{1} \gets 1,   \hspace{1cm} r_{2} \gets 1$ 
\State $Z^{k}_{1} \gets X - L^{k} - S^{k}, \hspace{1cm} Z^{k}_{2} \gets W^{k} - L^{k}$  \Comment{Initialize dual variables }
\State $P^{k}_{1} \gets \|L^{k}\|_{*}, \hspace{1cm} P^{k}_{2} \gets \lambda\|S^{k}\|_{1}, \hspace{1cm} P^{k}_{3} \gets \gamma\tr(L^{k}\Phi {L^{k}}^{T})$ \Comment{Initialize primal objective}
\While{$\frac{\|P^{k+1}_{1} - P^{k}_{1}\|^{2}_{F} }{\|P^{k}_{1}\|^{2}_{F}} > \epsilon \hspace{0.2cm} \& \hspace{0.2cm}  \frac{\|P^{k+1}_{2} - P^{k}_{2}\|^{2}_{F} }{\|P^{k}_{2}\|^{2}_{F}} > \epsilon \hspace{0.2cm}  \& \hspace{0.2cm}  \frac{\|P^{k+1}_{3} - P^{k}_{3}\|^{2}_{F} }{\|P^{k}_{3}\|^{2}_{F}} > \epsilon \hspace{0.2cm} \& \hspace{0.2cm}   \frac{\|Z^{k+1}_{1} - Z^{k}_{1}\|^{2}_{F} }{\|Z^{k}_{1}\|^{2}_{F}} > \epsilon \hspace{0.2cm} \& \hspace{0.2cm}  \frac{\|Z^{k+1}_{2} - Z^{k}_{2}\|^{2}_{F} }{\|Z^{k}_{2}\|^{2}_{F}} > \epsilon$}
\State Update $L^{k+1}$ using eq. 5
\State Update $S^{k+1}$ using eq. 6
\State Update $W^{k+1}$ using eq. 7
\State Update $Z^{k+1}_{1}$ using eq. 2
\State Update $Z^{k+1}_{2}$ using eq. 3
\State Update $P^{k+1}_{1},P^{k+1}_{2}$ and $P^{k+1}_{3}$ using step 6.
\State $k \gets k + 1$
\EndWhile\label{euclidendwhile}
\State \textbf{return} $L^{k+1}, S^{k+1}$
\EndProcedure
\end{algorithmic}
\end{algorithm}

\FloatBarrier
\subsection{Convergence Analysis \& Computational Complexity}\label{sec:convergence}
Algorithm 1 is a special case of {Alternating Directions} method \cite{yuan2009sparse}, \cite{boyd2010distributed}. These methods are a subset of  more general class of methods known as {Augmented Lagrange Multiplier} methods. The convergence of these algorithms is well-studied \cite{boyd2010distributed}, \cite{lions1979splitting}, \cite{kontogiorgis1998variable}. This algorithm has been reported to perform reasonably well on a wide range of problems and small number of iterations are enough to achieve a good accuracy \cite{candes2011robust}.  The complexity of nuclear norm proximal computation is $O(np^{2}+p^{3})$ for $n>p$ and the computational complexity of the Conjugate Gradient method for updating $W$ is $O(np)$, per iteration. Thus, the dominant cost of each iteration corresponds to the computation of nuclear proximal operator. Our future work will be dedicated to reduce this cost by utilizing a partial SVD or an approximate SVD, as suggested in \cite{goldfarb2011convergence}. Further improvements can be made by using randomized algorithms for SVD \cite{witten2013randomized} and exploiting the parallel processing capabilities \cite{lucas2014parallel}. Please refer to Section~\ref{sec:time} for a detailed comparison of computation time of this algorithm with other models considered in this work.

\clearpage
\subsection{Properties of Various Models \& Datasets}\label{sec:properties}

\begin{table}[htbp]
\footnotesize
\caption{A summary of the datasets used for the evaluation of various models. As NMF \& GNMF require non-negative data so they are not evaluated for USPS, MFeat and BCI datasets.}
\centering
\resizebox{0.55\textwidth}{!}{\begin{tabular}[t]{| c | c | c | c | }\hline
\textbf{Image /} & \textbf{Data}     & \textbf{Datasets} & \textbf{Evaluated} \\
 \textbf{ non-image}     &  \textbf{Type}   &      & \textbf{Models}\\\hline
Image     & Faces     & CMU PIE (no pose changes)    & all  \\\cline{3-3}
                &    &  ORL (pose changes)  &  \\\cline{3-3}
                &    &  YALE (facial expressions)  &  \\\cline{2-3}
                &  Objects   & COIL20 (pose changes)   &  \\\cline{2-3}
                &  Digits        & MNIST  (rotation)    & \\\cline{3-4}
                &                  &     USPS        &  \\\cline{1-3}
non-    &    features        &    BCI (Brain Computer Interface)         & all except \\\cline{3-3}
   image              &               &  MFeat (handwritten numerals)        &   NMF \& GNMF \\\hline           
\end{tabular}}
\label{tab:datasets}
\end{table}
        
      \begin{figure*}[htbp]
    \centering
        \centering
        \includegraphics[width=0.65\textwidth]{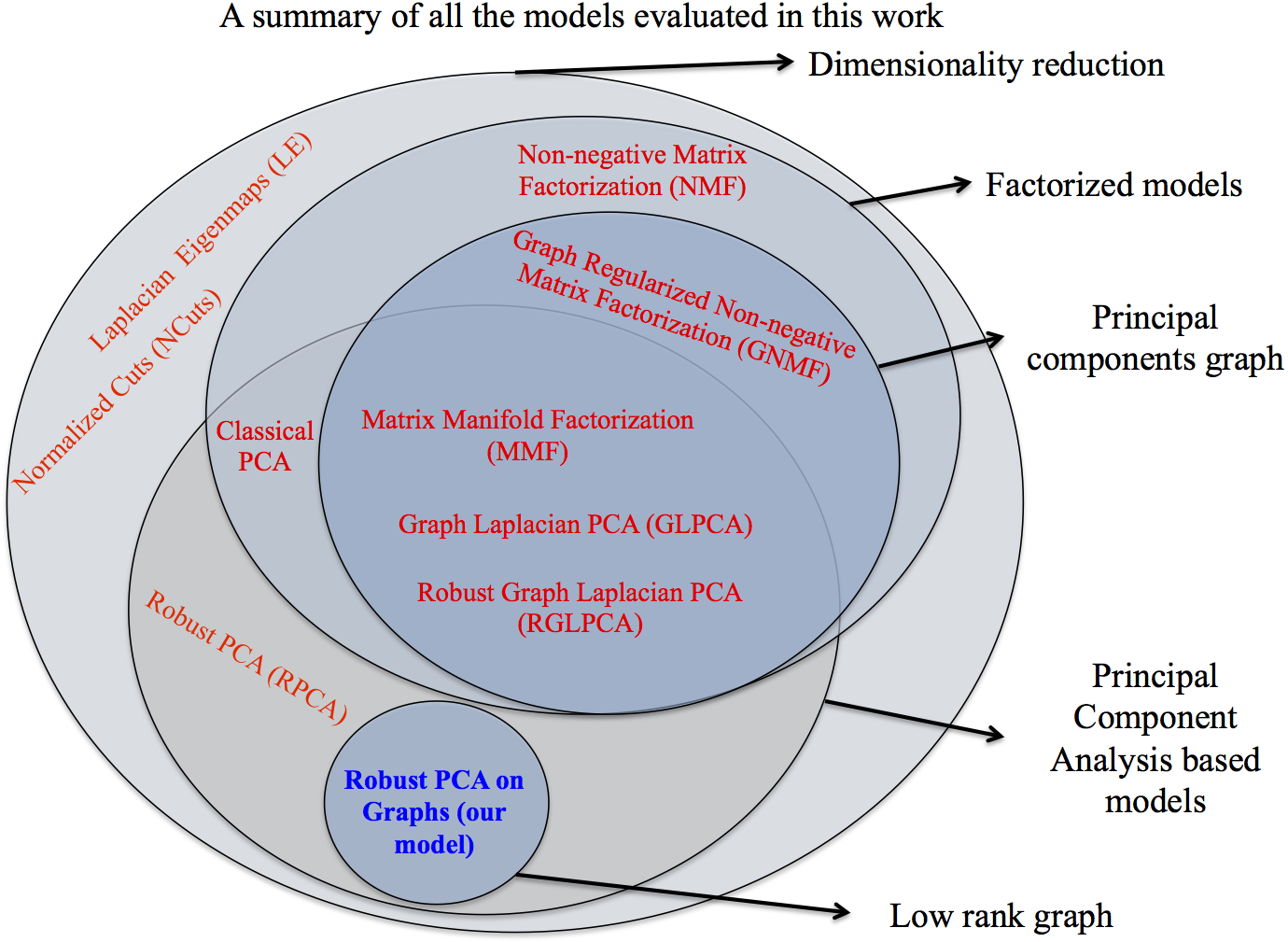}
         \caption{A venn diagram summarizing the properties of all the models evaluated in this work. }
        \label{fig:venn}
    \end{figure*}

\begin{figure}[!]
    \centering
        \centering
        \includegraphics[width=0.45\textwidth]{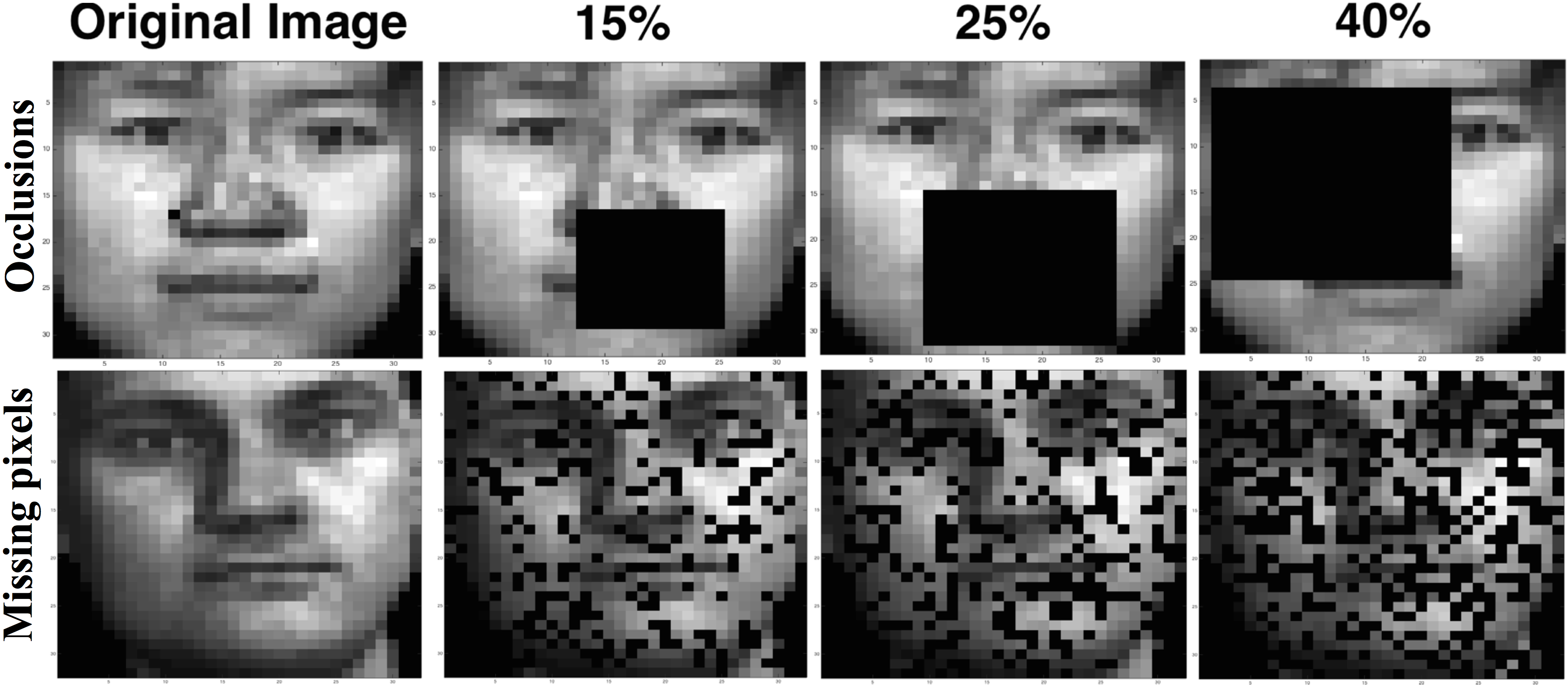}
         \caption{Sample images from CMU PIE dataset corrupted with occlusions (first row) and missing pixels (2nd row). }
        \label{fig:corruptions}
    \end{figure}

\clearpage
\FloatBarrier

   \subsection{Evaluation Scheme and Parameter Selection for all Models}  \label{sec:param_sel}
   
   \FloatBarrier
  \begin{figure*}[htbp]
    \centering
        \centering
        \includegraphics[width=0.65\textwidth]{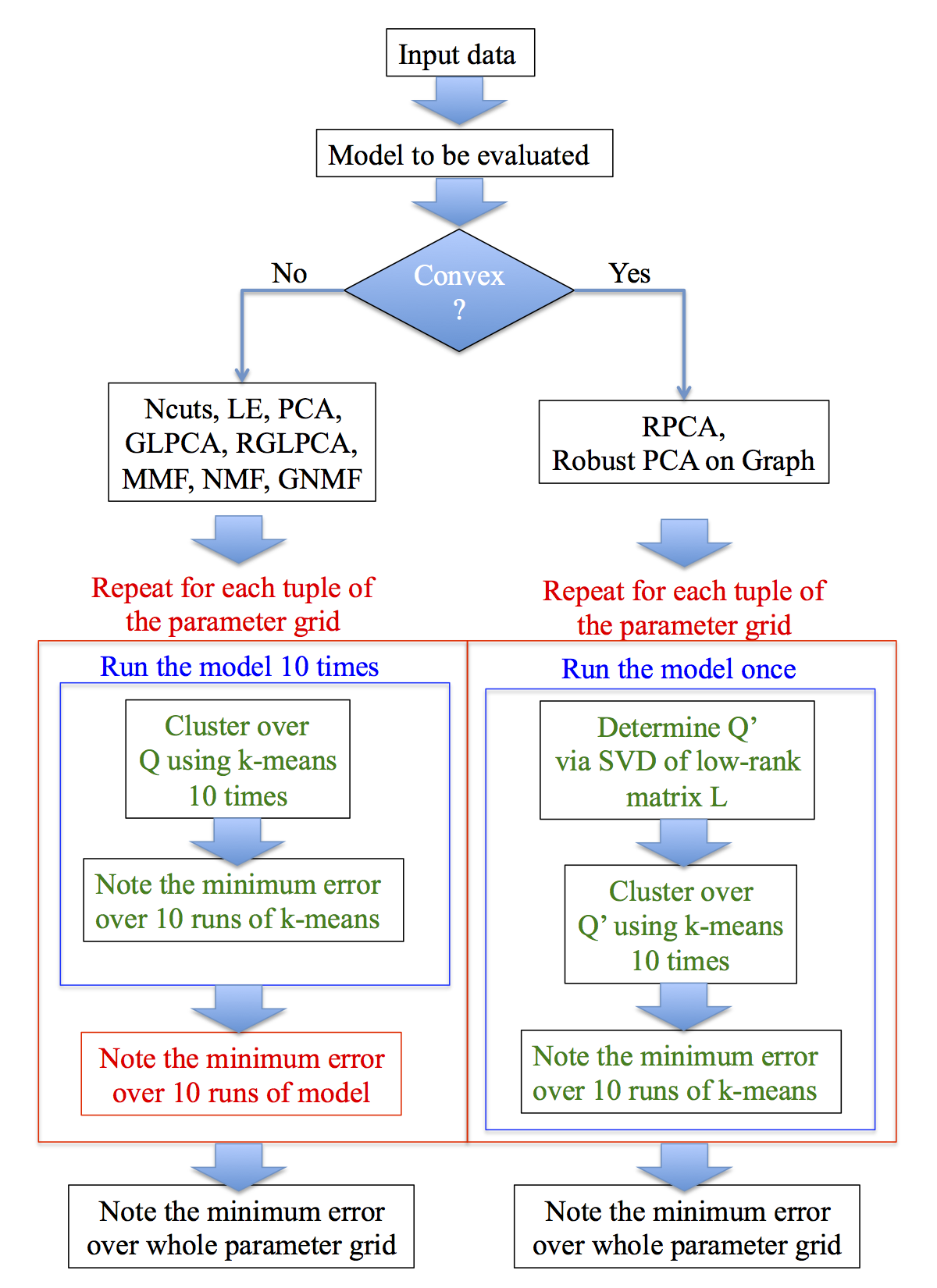}
         \caption{A flow chart describing the evaluation and parameter selection procedure for each of the models considered in this work. Each model has several parameters as described in Table~\ref{tab:models_param}. To perform a fair evaluation between all the models, they are run over the entire parameter range and the minimum error is reported. Further, the non-convex models are run 10 times (to determine a good local minimum)  for every tuple of the parameter range and the minimum error is reported. The k-means clustering procedure for each of the models is also evaluated 10 times to avoid the bias introduced by its non-convexity. }
        \label{fig:flow_clustering}
    \end{figure*}

\FloatBarrier       
\begin{sidewaystable*}[htbp]
\footnotesize
\caption{Range of parameter values for each of the models considered in this work. The clustering results reported in Tables~\ref{tab:results1},~\ref{tab:results1a} \&~\ref{tab:results2} correspond to the minimum error over this parameter range using the procedure of Fig.~\ref{fig:flow_clustering}. $X \in \mathbb{R}^{p\times n}$ is the data matrix, $U \in \mathbb{R}^{p\times d}$ and $Q \in \mathbb{R}^{d\times n}$ are the {principal directions} and {principal components} in a $d$ dimensional space (rank = $d$). $L = UQ$ is the {low-rank representation} and $S$ is the sparse matrix. $D$ is the degree matrix and $A$ is the adjacency matrix. $D-A$ is the unnormalized graph Laplacian and $\Phi = D^{-1/2}(D-A)D^{-1/2}$  is the normalized graph Laplacian. $\|\cdot\|_{F}$, $\|\cdot\|_{*}$ and $\|\cdot\|_{1}$ denote the Frobenius, nuclear and $l_{1}$ norms respectively.}
\centering
\resizebox{1.0\textwidth}{!}{\begin{tabular}[t]{| c | c | c | c | c | c | c |} \hline
   &   \textbf{Model}   & \textbf{Objective}  &   \textbf{Constraints}   & \textbf{Parameters}  & \textbf{Parameter Range} & \textbf{no. of runs for each}   \\
   &         &        &     &    &    & \textbf{parameter value}   \\\hline
   1& NCut  \cite{shi2000normalized} &  $  \min_{Q} \tr(Q\Phi Q^{T}) $  &  $ QQ^{T} = I $ &  & & \\\cline{1-4}
 2 & LE \cite{belkin2003laplacian} & $   \min_{Q} \tr(Q(D-A) Q^{T}) $   &   $ QDQ^{T} = I$   & $d$ &  $d\in \{2^{1},2^{2},\cdots, \min(n,p) \}$  &  10 \\\cline{1-4}
3 &  PCA    &  $ \min_{U,Q} \|X-UQ\|_{F}^{2}$  &   $U^{T}U  = I$ & & & (non-convex)  \\\hline
4  &  GLPCA  \cite{jiang2013graph}   & $\min_{U,Q} \|X-UQ\|_{F}^{2}  + \gamma \tr(Q\Phi Q^{T})$ &   & & $d\in \{2^{1},2^{2},\cdots, \min(n,p) \}$   &    \\\cline{1-3}
5  & RGLPCA  \cite{jiang2013graph} & $\min_{U,Q} \|X-UQ\|_{2,1}  + \gamma \tr(Q\Phi Q^{T})$ & $QQ^{T} = I $ &  $d,\gamma$ & $\gamma \implies \beta$ using transformation in  \cite{jiang2013graph} & 10  \\
   &         &       &      &  &$\beta \in \{0.1, 0.2, \cdots, 0.9\}$   &   (non-convex)  \\\hline
6  & MMF  \cite{zhang2013low}      &  $\min_{U,Q} \|X-UQ\|_{F}^{2}  + \gamma \tr(Q\Phi Q^{T})$ & $U^{T}U = I $  & $d,\gamma$ &  $d\in \{2^{1},2^{2},\cdots, \min(n,p) \}$   &  \\\cline{1-5}
7  & NMF  \cite{lee1999learning} & $ \min_{U,Q} \|X-UQ\|_{F}^{2}$   & $U \geq 0, Q \geq 0$  &  $d$ &   & 10 \\\cline{1-5}
8  & GNMF \cite{cai2011graph}   &  $\min_{U,Q} \|X-UQ\|_{F}^{2}  + \gamma \tr(Q\Phi Q^{T})$   & $U \geq 0, Q \geq 0$  & $d,\gamma$ &  $\gamma \in \{2^{-3},2^{-2},\cdots, 1000\}$   & (non-convex) \\\hline
9   & RPCA \cite{candes2011robust}  &  $\min_{L,S} \|L\|_{*} + \lambda\|S\|_{1}$  &   $X = L + S$  & $\lambda$  & $\lambda \in \{\frac{2^{-3}}{\sqrt{\min(n,p)}}:0.01:\frac{2^{3}}{\sqrt{\min(n,p)}}\}$  &  1  \\\cline{1-5}
10 & Our model     &   $\min_{L,S} \|L\|_{*} + \lambda\|S\|_{1} + \gamma\tr(L\Phi L^{T})$ & $X = L + S$  & $\lambda, \gamma$ & $\gamma \in \{2^{-3},2^{-2},\cdots, 1000\}$   &  (convex) \\\hline 
\end{tabular}}
\label{tab:models_param}
\end{sidewaystable*}

\clearpage
\FloatBarrier
\subsection{Adjacency Matrix Construction from Corrupted \& Uncorrupted Data}\label{sec:laplacian}
\FloatBarrier
  \begin{figure*}[htbp]
    \centering
        \centering
        \includegraphics[width=0.7\textwidth]{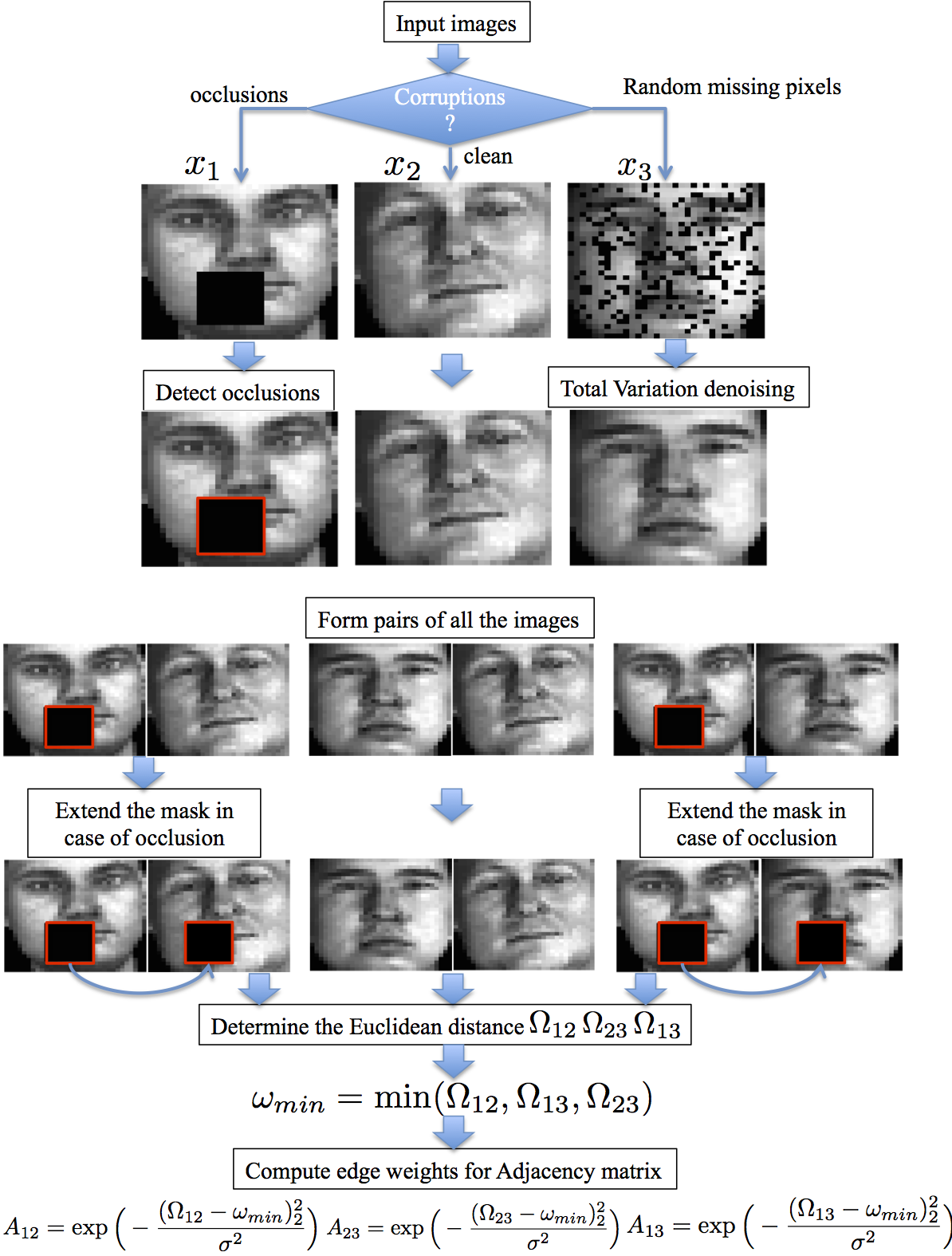}
         \caption{A procedure for the construction of Adjacency matrix $A$ from clean and corrupted samples $X$. We assume that the block occlusions can be detected and the mask can be extended to the other image of the pair to remove the effect of occlusion in the calculation of Euclidean distance. The case of random missing pixels can be handled by using inpainting techniques, such as Total Variation denoising. }
        \label{fig:laplacians}
    \end{figure*}
\clearpage
\FloatBarrier
\subsection{Clustering Results on all Databases}\label{sec:clustering_table1}
\FloatBarrier
\begin{table*}[htbp]
\footnotesize
\caption{A comparison of  clustering  error of our model with various dimensionality reduction models using the procedure of Fig.~\ref{fig:flow_clustering}. The image data sets include: 1) ORL 2) CMU PIE  3) YALE and 4) COIL20. The compared models are: 1) k-means 2) Normalized Cuts (NCuts) 3) Laplacian Eigenmaps (LE) \cite{belkin2003laplacian}  4) Standard Principal Component Analysis (PCA)  5) Graph Laplacian PCA (GLPCA) \cite{jiang2013graph} 6) Robust Graph Laplacian PCA (RGLPCA) \cite{jiang2013graph}  7) Manifold Regularized Matrix Factorization (MMF) \cite{zhang2013low}  8) Non-negative Matrix Factorization \cite{lee1999learning}  9) Graph Regularized Non-negative Matrix Factorization (GNMF) \cite{cai2011graph}  and 10) Robust PCA (RPCA) \cite{candes2011robust}. Two types of full and partial corruptions were introduced in the data: 1) Block occlusions and 2) Random missing values.  The best results are highlighted in bold. This table provides the numerical errors for the bar plots shown in Fig.~\ref{fig:images_error}.}
\centering
\resizebox{0.9\textwidth}{!}{\begin{tabular}[t]{| c || c || c || c | c || c | c | c || c | c | c | c |}\hline
\textbf{Data Set} & \textbf{Model}   & \textbf{No} & \multicolumn{2}{c ||}{\textbf{Sample specific corruptions}}  & \multicolumn{7}{c |}{\textbf{Full Corruptions}} \\\cline{4-12}
                           &                         &        \textbf{Corruptions}  & \textbf{Occlusions} & \textbf{Missing values} & \multicolumn{3}{c |}{\textbf{Occlusions}} & \multicolumn{4}{c |}{\textbf{Missing}} \\
                           &                         &                                      &  \textbf{(25\% of data)}  & \textbf{(25\% of data)} &  \multicolumn{3}{c |}{\textbf{(\% of image size)}} & \multicolumn{4}{c |}{\textbf{(\% of image pixels)}} \\\cline{4-12}
                           &       &             &  \textbf{25\% image size}    & \textbf{25\% image pixels}  & \textbf{15\%} & \textbf{25\%}   & \textbf{40\%}  & \textbf{15\%}  & \textbf{25\%} & \textbf{35\%}  & \textbf{50\%} \\\hline
 CMU PIE \footnotemark[1]           &   k-means &  72.2 & 72.5 & 73.4 &  78.5 &  79.4 & 81.0   & 72.3   & 76.1 & 75.7  & 80.3 \\\cline{2-12}
 (faces)                 &  NCuts   & 41.7 & 36.0 & 40.0 & 42.0 & 40.0 & 44.0 & 35.0  & 41.3  & 40.7 & 46.7 \\\cline{2-12}
                  & LE   & 41.7 & 40.0 & 43.3 & 49.3 & 54.0 & 51.3 & 46.3  & 46.0 & 48.3  & 55.0 \\\cline{2-12}
                  & PCA & 13.0 & 25.7  & 34.0 & 42.7 & 45.0 & 61.0 & 35.0 & 50.0 & 56.3  & 63.7 \\\cline{2-12}
                  & GLPCA   & 29.8 & 27.7 & 38.3 & 38.0  & 40.2 & 41.3 & 39.3 & 37.5 & 42.7 & 48.2 \\\cline{2-12}
                  & RGLPCA  & 28.0 & 28.3 & 31.7 & 35.0 & 41.3  & 39.0  & 34.3 & 33.7  & 28.7  & 42.0 \\\cline{2-12}
                  & MMF  & 53.7 & 50.0 & 55.0 & 54.0 & 58.0 & 60.7  & 59.7  & 60.3  & 60.3  & 62.7 \\\cline{2-12}
                  & NMF & 49.0 & 68.0 &  70.3 & 76.0  & 80.0 & 79.3 & 49.0 & 76.0 & 80.0 & 79.3 \\\cline{2-12}
                  & GNMF & 52.7 &  66.0  & 68.3 & 71.3 & 75.0 & 75.3 & 52.7 & 71.3 & 75.0 & 75.3 \\\cline{2-12}
                  & RPCA & 3.3 & \textbf{4.3}  & 11.0 & 6.7 & 13.3 & 23.3  & 9.7  & 8.3 & 17.0 & 37.3 \\\cline{2-12}
                  & Our model  & \textbf{0.0} & 4.7  & \textbf{6.9} & \textbf{5.0}  & \textbf{7.7}  & \textbf{17.7}  &  \textbf{7.7}  & \textbf{7.5} & \textbf{15.7}  & \textbf{34.0} \\\hline\hline
ORL\footnotemark[2]            &   k-means &   35.4  & 57.6  &  50.8 &  64.3 & 68.4 & 77.5  & 36.0 & 42.4  & 49.4 & 57.1 \\\cline{2-12}
 (faces)                 &  NCuts  & 47.0 &  35.7 &  43.0  &   42.3 & 53.0  & 61.0 & 49.7  & 45.7  & 42.7  & 46.7 \\\cline{2-12}
                  & LE   & 45.3 &  38.0  &   34.7 &  39.0   & 45.3  & 54.3 & 42.7  & 45.0 & 44.0  & 42.7 \\\cline{2-12}
                  & PCA   & 28.0  & 46.0  &  37.3 &  53.7  & 65.0  & 72.6  & 34.3 & 39.0 & 40.3 & 40.0 \\\cline{2-12}
                  & GLPCA   & 27.6  & 28.3  &  29.7 & 33.6  & 34.1  & 44.3  & 30.3 & 27.9 & 27.6 & 33.3 \\\cline{2-12}
                  & RGLPCA  & 28.3  & 28.7  & 29.0  & 28.3  & 36.3  & 37.3  & 26.0 & 29.3 & 26.7  & 26.3 \\\cline{2-12}
                  & MMF  & 20.3 &  29.7 & 24.3  &  30.0 & 34.7  & 38.3  & 24.7  & 28.3  & 28.3  & 27.3 \\\cline{2-12}
                  & NMF  & 29.0  & 31.7  & 27.7  & 78.0 & 79.0  & 81.3  & 29.7  & 39.3  & 50.3  & 53.3 \\\cline{2-12}
                  & GNMF & 22.7 &  29.0  &  25.3 &  35.0 & 37.0 & 39.3 & 25.3 & 26.7 & 28.0  & 26.7 \\\cline{2-12}
                  & RPCA & 18.6  & 27.7  &  20.3 &  26.0  &  35.0  & 71.4 & 24.7 & 26.0 & 27.0 & 36.0 \\\cline{2-12}
                  & Our model  &  \textbf{15.7}  & \textbf{20.0}  & \textbf{14.3}  & \textbf{23.7} & \textbf{24.7} & \textbf{31.7}  & \textbf{20.0} &  \textbf{18.3}  & \textbf{20.3}  & \textbf{21.0}  \\\hline\hline
YALE   \footnotemark[3]        &   k-means &   53.4  & 54.9  & 64.5  &  71.6 & 73.4 &  73.1  &   54.3    &   56.4  &  58.9    & 63.5    \\\cline{2-12}
 (faces)                 &  NCuts  & 54.5  &  57.6 &  64.4 &  66.7  & 64.2  & 68.5 &   56.4  & 56.9  & 61.2  & 61.8 \\\cline{2-12}
                  & LE   & 57.6 & 57.0   &  61.8 & 63.6   & 68.5  & 69.1 &  57.8  &  59.4  &  60.6   &  66.7 \\\cline{2-12}
                  & PCA   & 53.9  & 49.7  & 58.8  &  61.8  & 66.1  & 70.9   & 55.7  &  55.1 &  62.4  & 61.2 \\\cline{2-12}
                  & GLPCA   & 49.1  &  50.9 & 53.9  & 54.5  & 61.2  & 58.2 & 50.9 & 52.7   &   48.5  &  56.9 \\\cline{2-12}
                  & RGLPCA  & 48.5  & 50.0  &  50.9 & 54.5  & 55.2  & 58.8 & 50.3 &  53.9 &  49.7 & 50.9 \\\cline{2-12}
                  & MMF  & 38.8 & 37.6   & 46.7  &  55.7 & 55.2  & 53.9  &  38.2  & 38.2   &  44.2 & 49.1 \\\cline{2-12}
                  & NMF  &  63.0 &  63.6  &  70.3 & 72.1 &  71.5 & 72.1 &  64.8  &  61.2  &  61.8  & 66.7 \\\cline{2-12}
                  & GNMF & 56.9  & 55.8  &  57.6 & 60.6  & 64.8 &  60.6 & 56.9 &  56.9  &  58.2   &  59.4 \\\cline{2-12}
                  & RPCA & 39.4  &  40.6 & 45.5  &  61.8  &  67.9  & 63.0  & 42.4 & 39.4  & 43.6  &  63.6 \\\cline{2-12}
                  & Our model  &  \textbf{35.1}  & \textbf{35.8}  & \textbf{40.0}   & \textbf{43.6} & \textbf{46.1} & \textbf{50.9} & \textbf{35.2} &  \textbf{35.8} & \textbf{39.4}  & \textbf{50.3} \\\hline\hline
COIL20 \footnotemark[4]           &   k-means &  32.0 & 55.2 & 30.2 & 34.9 & 45.8 & 57.8 & 36.3 & 34.8 & 37.8  & 42.8 \\\cline{2-12}
 (objects)                 &  NCuts  & 44.5 & 30.5 & 39.5 & 38.5 & 38.5  & 43.5  & 38.0  & 50.0  & 48.0  & 47.5 \\\cline{2-12}
                  & LE   & 38.0 &  31.5 & 27.0 & 34.5 & 36.5  & 40.0  & 31.0  & 37.5  & 40.0  & 39.0 \\\cline{2-12}
                  & PCA  & 30.5 & 43.5 & 28.5 & 28.0 & 37.0  & 50.0 & 24.0 & 30.0  & 33.5  & 31.0 \\\cline{2-12}
                  & GLPCA  & 20.5 &  22.0 & 17.0 & 25.0  & 25.5  & 25.3 & 21.7 & 21.5 & 21.5 & 23.3 \\\cline{2-12}
                  & RGLPCA  & 18.5 & 23.5 & 17.5 & 20.5 & 22.0  & 23.5  & 22.0  & 20.0  & 22.0  & 21.0 \\\cline{2-12}
                  & MMF  & 18.0 & 19.0 & 11.5 & 19.0  & 18.0  & 25.0 & 19.0  & 18.0 & 19.0 & 18.5 \\\cline{2-12}
                  & NMF  & 20.0 & 30.0 &19.0 & 32.5  & 47.0 & 60.5  & 17.5  & 26.0  & 24.0 & 26.0 \\\cline{2-12}
                  & GNMF & \textbf{15.5} & 18.5 & 9.5 & 20.0 & 19.5 & 30.0  & 14.5  & 13.5 & \textbf{11.0} & 17.0 \\\cline{2-12}
                  & RPCA & 19.0 & 33.0 & 13.5 & 19.0 & 28.0 & 50.5 & 15.0 & 16.0 & 16.5 & 27.0 \\\cline{2-12}
                  & Our model & \textbf{15.5} & \textbf{18.0} & \textbf{9.0} & \textbf{18.0} & \textbf{17.5} & \textbf{19.5}  & \textbf{8.5}  & \textbf{12.0}  & 16.5  & \textbf{15.0} \\\hline\hline
\end{tabular}}
\label{tab:results1}
\end{table*}

\begin{table*}[htbp]
\footnotesize
\caption{A comparison of  clustering  error of our model with various dimensionality reduction models. The image data sets include: 1) MNIST and 2) USPS. The compared models are: 1) k-means 2) Normalized Cuts (NCuts) \cite{shi2000normalized} 3) Laplacian Eigenmaps (LE) \cite{belkin2003laplacian}  4) Standard Principal Component Analysis (PCA)  5) Graph Laplacian PCA (GLPCA) \cite{jiang2013graph} 6) Robust Graph Laplacian PCA (RGLPCA) \cite{jiang2013graph}  7) Manifold Regularized Matrix Factorization (MMF) \cite{zhang2013low}  8) Non-negative Matrix Factorization \cite{lee1999learning}  9) Graph Regularized Non-negative Matrix Factorization (GNMF) \cite{cai2011graph}  and 10) Robust PCA (RPCA) \cite{candes2011robust}. Two types of full and partial corruptions were introduced in the data: 1) Block occlusions and 2) Random missing values. The best results are highlighted in bold.}
\centering
\resizebox{1.0\textwidth}{!}{\begin{tabular}[t]{| c || c || c || c | c || c | c | c || c | c | c | c |}\hline
\textbf{Data Set} & \textbf{Model}   & \textbf{No} & \multicolumn{2}{c ||}{\textbf{Sample specific corruptions}}  & \multicolumn{7}{c |}{\textbf{Full Corruptions}} \\\cline{4-12}
                           &                         &        \textbf{Corruptions}  & \textbf{Occlusions} & \textbf{Missing values} & \multicolumn{3}{c |}{\textbf{Occlusions}} & \multicolumn{4}{c |}{\textbf{Missing}} \\
                           &                         &                                      &  \textbf{(25\% of data)}  & \textbf{(25\% of data)} &  \multicolumn{3}{c |}{\textbf{(\% of image size)}} & \multicolumn{4}{c |}{\textbf{(\% of image pixels)}} \\\cline{4-12}
                           &       &             &  \textbf{25\% image size}    & \textbf{25\% image pixels}  & \textbf{15\%} & \textbf{25\%}   & \textbf{40\%}  & \textbf{15\%}  & \textbf{25\%} & \textbf{35\%}  & \textbf{50\%} \\\hline
MNIST\footnotemark[5]            &   k-means &   62.2 & 71.2  & 59.6 & 68.2 & 79.2 & 82.5  & 62.0  & 60.5  & 60.6  & 65.9 \\\cline{2-12}
(digits)                  &  NCuts  &  69.3 &  86.7 & 59.0 & 85.3  & 87.3  & 86.0  & 77.7  & 78.7  & 83.7  & 86.7 \\\cline{2-12}
                 & LE   &  71.0 & 88.0 & 55.3 & 88.0  & 89.3  & 89.0  & 68.7  & 55.3  & 86.3  & 89.3 \\\cline{2-12}
                  & PCA   & 52.0 & 60.0 & 51.3  & 68.3 & 77.0 & 77.3 & 46.0  & 57.3  & 61.0  & 57.0 \\\cline{2-12}
                 & GLPCA   & 49.3 & 63.7 & 44.7 &  59.3 & 71.7  & 79.0  & 45.3  & 51.7 & 56.3 & 57.1 \\\cline{2-12}
                  & RGLPCA  & 46.3 & 64.7 & 51.0  & 55.3 & 65.3  & 73.7  & 52.7  & 53.0 & 53.7 & 44.3 \\\cline{2-12}
                  & MMF  &  45.7 & 65.7 & 45.0 &  52.7  & 65.3  & 77.3  & 52.7 & 44.7  & 53.3  & 55.7 \\\cline{2-12}
                & NMF  & 46.3 & 67.3 & 33.7 & 60.0  & 72.3  & 72.0 & 52.0  & 47.7  & 49.7 & 47.0 \\\cline{2-12}
               & GNMF & 49.7 & 72.7 & 35.3 & 51.7  & 71.0  & 83.7 & 50.7 & 42.7 & 44.7 & 40.7   \\\cline{2-12}
                                 & RPCA & 39.0 & 53.3  & 43.0 & 64.7  & 73.7  & 78.0  & 44.7 & 44.7 & 52.7  & 70.7   \\\cline{2-12}
                 & Our model  &  \textbf{31.7}      & \textbf{46.0}  & \textbf{32.3} & \textbf{53.0} & \textbf{62.7}  & \textbf{69.7}  & \textbf{29.7}  & \textbf{33.7}  & \textbf{35.3}  & \textbf{37.7} \\\hline\hline
USPS            &   k-means & 51.0 & 45.3 & 54.0 & 55.4 & 66.2 & 73.8 & 41.2 & 41.7 & 40.1  & 38.9 \\\cline{2-12}
(digits)                  &  NCuts  & 54.0 & 49.3 & 61.3 & 68.3 & 72.7  & 75.7 &  53.0  &  47.7  & 55.7 & 55.3 \\\cline{2-12}
                  & LE   & 64.7  & 52.7 & 67.7 & 74.3 &  81.7 & 83.0 & 65.0 &  71.7  & 66.0  & 63.3 \\\cline{2-12}
                  & PCA & 49.7  & 42.0 &  50.3 & 55.0 & 66.0 & 71.7 & 43.3   & 36.0 & 39.3  & 48.0 \\\cline{2-12}
                  & GLPCA   & 45.8  & 26.7 & 45.0 & 46.9  &  56.7 & 68.9  & 39.7   & 38.3 & 33.5 & 32.5 \\\cline{2-12}
                  & RGLPCA  & 42.0  & 33.3 & 46.7 & 45.0 & 59.0  & 63.7 & 33.3  & 26.7  & 35.7 & 37.0 \\\cline{2-12}
                  & MMF  & 33.0 & 22.3 &  40.7 & \textbf{37.7} & 50.3  & 62.6 & \textbf{21.5}  &  21.6  & 20.7  & \textbf{20.7} \\\cline{2-12}
                  & RPCA & 28.3 & 29.0 & 46.7  & 56.0  & 67.7 & 69.7 & 29.0  &  32.3 & 32.6  & 38.0 \\\cline{2-12}
                  & Our model  &  \textbf{21.3} & \textbf{21.7}  & \textbf{35.3}  & 42.0  &  \textbf{48.0} & \textbf{56.0} & \textbf{21.5}    &  \textbf{21.3}  & \textbf{20.5}     & 27.3  \\\hline                                                                         
\end{tabular}}
\label{tab:results1a}
\end{table*}


 \begin{table*}[htbp]
\footnotesize
\caption{A comparison of clustering error of PCA models and simple k-means for MFeat and BCI data sets. Each of the data sets was corrupted with two types of outliers: 1) Block occlusions and 2) Missing values. Block occlusions in non-image databases correspond to an unrealistic assumption so such experiments were no performed for these datasets. Furthermore, NMF and GNMF were not evaluated because they require non-negative data whereas these datasets are negative as well. The best results are highlighted in bold.}
\centering
\resizebox{0.7\textwidth}{!}{\begin{tabular}[t]{| c | c | c | c | c | c | c | c |}\hline
\textbf{Data} & \textbf{Model}   & \textbf{No} & {\textbf{Sample specific}}  & \multicolumn{4}{c |}{\textbf{Full Corruptions}} \\\cline{5-8}
        \textbf{Set}                   &                         &        \textbf{Corruptions}  & \textbf{Missing values } & \multicolumn{4}{c |}{\textbf{Missing values}} \\
                           &                         &                                &  \textbf{(25\% of data)}  & \multicolumn{4}{c |}{\textbf{(\% features per sample)}} \\\cline{5-8}    
                           &             &                &   \textbf{25\% features per sample}         &          \textbf{15\%}   & \textbf{25\%}    & \textbf{35\%}    & \textbf{50\%} \\\hline
{MFeat}   \footnotemark[6]            &  k-means  & 32.4   & 30.8  & 25.5  & 29.9  & 32.2  & 33.9  \\\cline{2-8}
                            & NCuts      & 39.8   & 47.0    &  26.3  & 28.3   & 33.5  & 37.8   \\\cline{2-8}
                            & LE           &  38.0   &   47.3  &   35.3  & 33.0   & 33.5  & 50.8  \\\cline{2-8}
                            &  PCA   & 7.3  &  7.3  &  10.0 & 11.8   & 9.0   & 13.8   \\\cline{2-8}
                            & GLPCA & 6.0   &  12.5    &   6.0   & 7.8   & 8.0   & 10.0   \\\cline{2-8}
                            & RGLPCA & 9.5  & 15.0   & 15.0 &  17.3  & 8.3  & 14.8   \\\cline{2-8}
                            & MMF  & 7.3  & 6.3  &  6.0    & 5.3   & 7.0   & 7.0    \\\cline{2-8}
                            & RPCA & 5.0 &  4.3   &  3.8   & 6.3  & 9.5  & 15.0   \\\cline{2-8}
                            & Our model   & \textbf{3.3}   &   \textbf{2.0}  &  \textbf{3.5}        & \textbf{4.3}   & \textbf{6.8} & \textbf{7.0}     \\\hline\hline
                          {BCI}   \footnotemark[7]            &  k-means  & 47.8   & 47.2  & 47.8  & 47.8  & 48.0  & 48.2  \\\cline{2-8}
                            & NCuts      & 47.0   &  47.0   &  47.3  & 47.5   & 47.3  & 48.3   \\\cline{2-8}
                            & LE           &  46.5   & 49.7    &   49.8 & 49.8  & 49.8  & 45.8  \\\cline{2-8}
                            &  PCA   & 45.3  &  45.2  &  42.0 & 43.0   & 43.0   & 45.5   \\\cline{2-8}
                            & GLPCA & 46.5   &  46.0    &   45.8   & 45.8   & 46.5   & 45.0   \\\cline{2-8}
                            & RGLPCA & 45.5  & 43.7   & 46.5 &  44.3 & 43.8  & 42.8   \\\cline{2-8}
                            & MMF  & 47.25  &  47.2 &  47.0    & 47.5   & 47.3   & 48.3    \\\cline{2-8}
                            & RPCA & \textbf{39.8} & 43.0    &  40.0   & 42.8  & 43.5  & 43.0   \\\cline{2-8}
                            & Our model   & 40.3   &  \textbf{37.7}   &   \textbf{39.0}        & \textbf{42.3}   & \textbf{42.0} & \textbf{41.0}     \\\hline\hline
\end{tabular}}
\label{tab:results2}
\end{table*}

\clearpage
\FloatBarrier
\subsection{A Comparison of the Principal Directions $U$ Learned with Low-Rank and Principal Components Graph}\label{sec:graph}
\FloatBarrier

      \begin{figure*}[htbp]
    \centering
        \centering
        \includegraphics[width=0.6\textwidth]{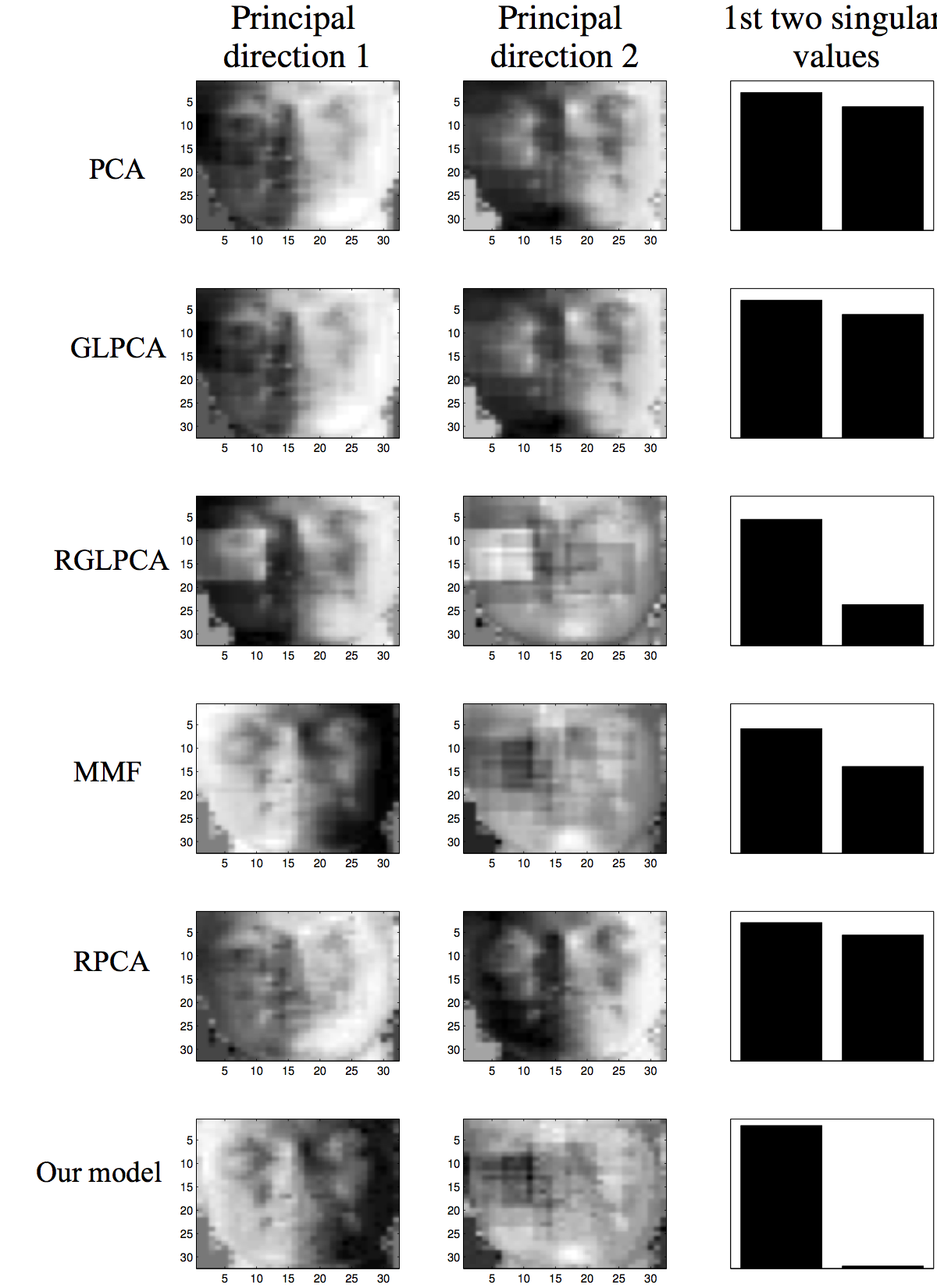}
         \caption{A comparison of the {principal directions $U$} learned with {low-rank graph $\tr(L\Phi L^{T})$} and {principal components graph regularization $\tr(Q\Phi Q^{T})$} for the CMU PIE dataset. Each row shows the $1^{st}$ two principal directions $U$ and the corresponding singular values for each of the PCA models evaluated in this work. For the factorized models (PCA, GLPCA, RGLPCA and MMF) the principal directions  $U$ are explicitly learned. For RPCA and our model, $U$ is obtained by singular value decomposition of the low-rank matrix $L$, i.e. $L = U\Sigma Q$. The {low-rank graph} for our model clearly helps in learning a corruption free $U$. Even though the $2^{nd}$ principal direction for our model has a large effect of occlusion, we note that the corresponding singular value ($2^{nd}$ bar) is much smaller as compared to the $1^{st}$. For all the other models, each of the two principal directions  are affected by occlusions and the corresponding singular values are significant. {\textbf{This explains why a {low-rank graph} helps in an enhanced low-rank recovery and a lower clustering error in low dimensional space than the {principal components graph}}}.} 
        \label{fig:basisvectors}
    \end{figure*}

\clearpage
\FloatBarrier
\subsection{Additional Low-Rank Recovery Results on CMU PIE Dataset}\label{sec:lowrank}
\FloatBarrier

      \begin{figure*}[htbp]
    \centering
        \centering
       \includegraphics[width=0.95\textwidth]{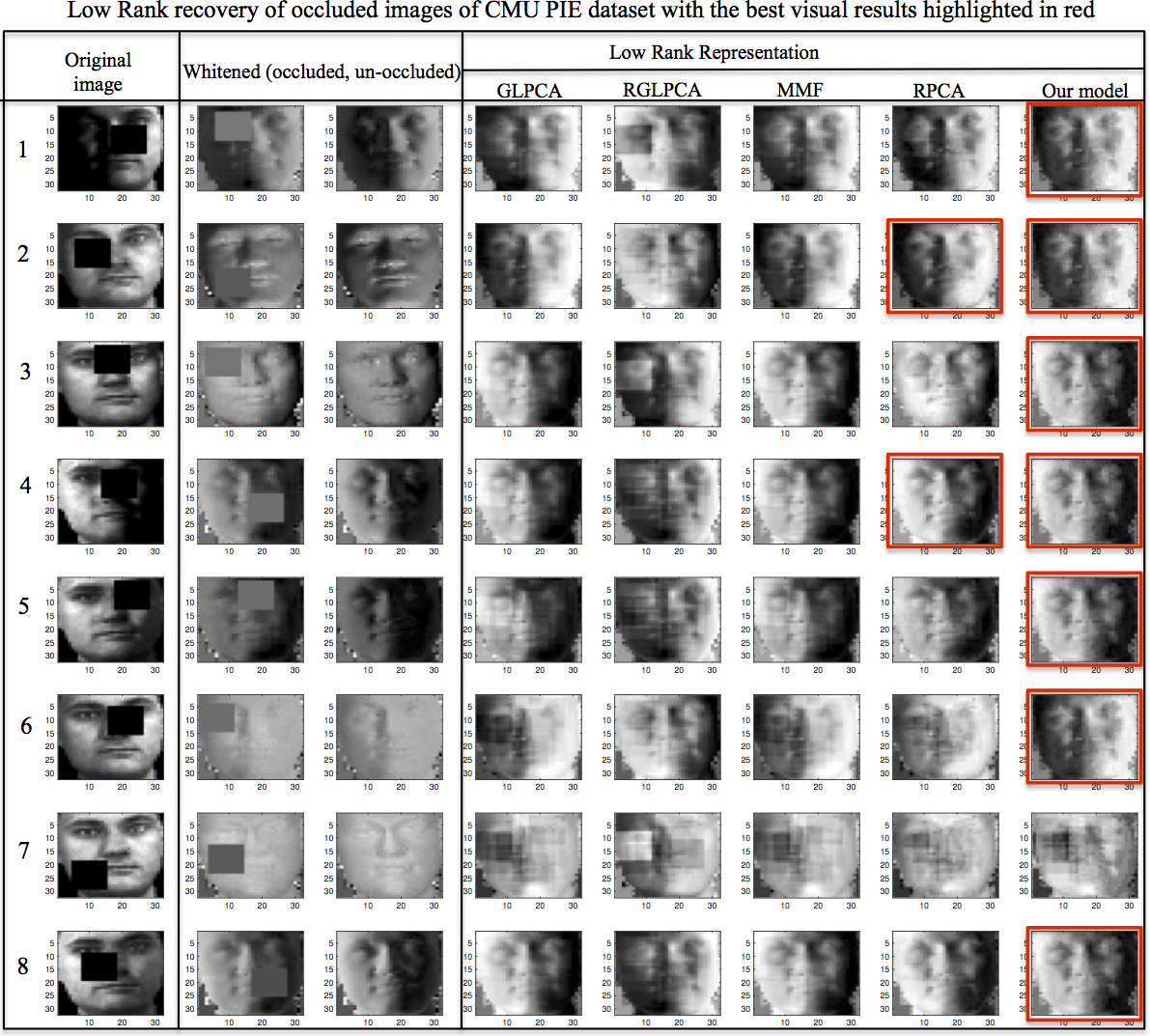}
         \caption{A comparison of the clean low-rank recovered images of the CMU PIE data set corresponding to each of the PCA models considered in this work. For this experiment, the images of one person are corrupted by block occlusions occupying 10\% of the image. Each row corresponds to a different image of the same person occluded at a random position. $1^{st}$ figure in each row shows the actual occluded image. $2^{nd}$ and $3^{rd}$ figures show the whitened occluded and un-occluded images. Since PCA requires whitening, the recovered low-rank images in figures 4 to 8 using GLPCA, RGLPCA, MMF, RPCA and our model resemble the un-occluded whitened image. The best recovered representations (via a visual inspection) are highlighted with a red box. }
        \label{fig:imagesLSfull}
    \end{figure*}

\clearpage
\FloatBarrier
\subsection{Additional Results for Background Separation from Videos}\label{sec:videos}
\FloatBarrier
\begin{figure*}[htbp]
    \centering
     \begin{subfigure}[b]{1.0\textwidth}
        \centering
        \includegraphics[width=1.0\textwidth]{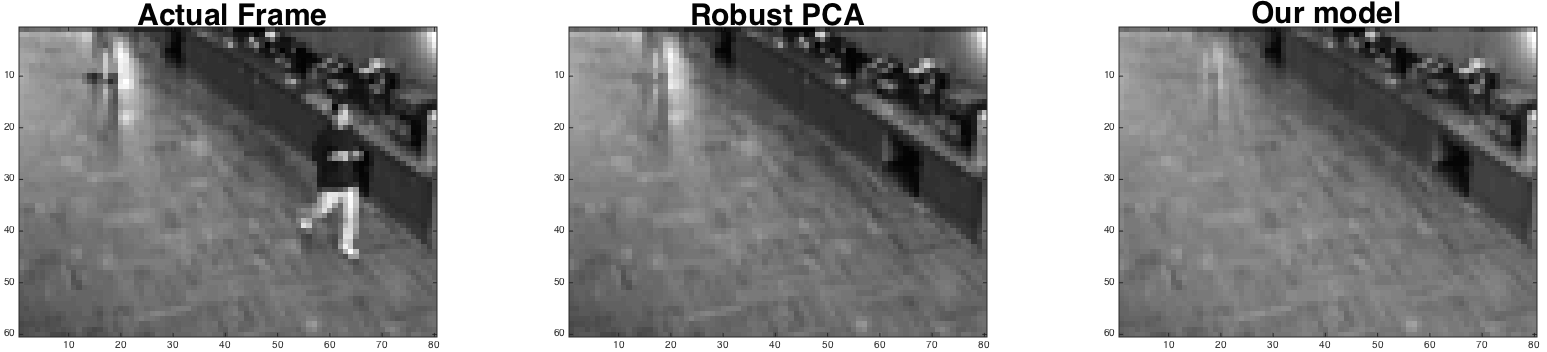}
        \caption{Restaurant food counter. The lightning effect (not static) is less visible in right figure than the middle one. }
    \end{subfigure}
    \hfill
    \begin{subfigure}[b]{1.0\textwidth}
        \centering
        \includegraphics[width=1.0\textwidth]{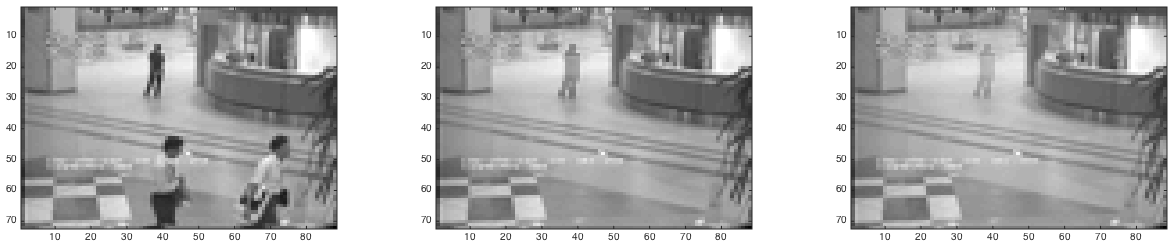}
        \caption{Airport hallway. The moving person (not a part of static background) is less visible in the right figure than the middle one.}
    \end{subfigure}
    \hfill
    \begin{subfigure}[b]{1.0\textwidth}
        \centering
        \includegraphics[width=1.0\textwidth]{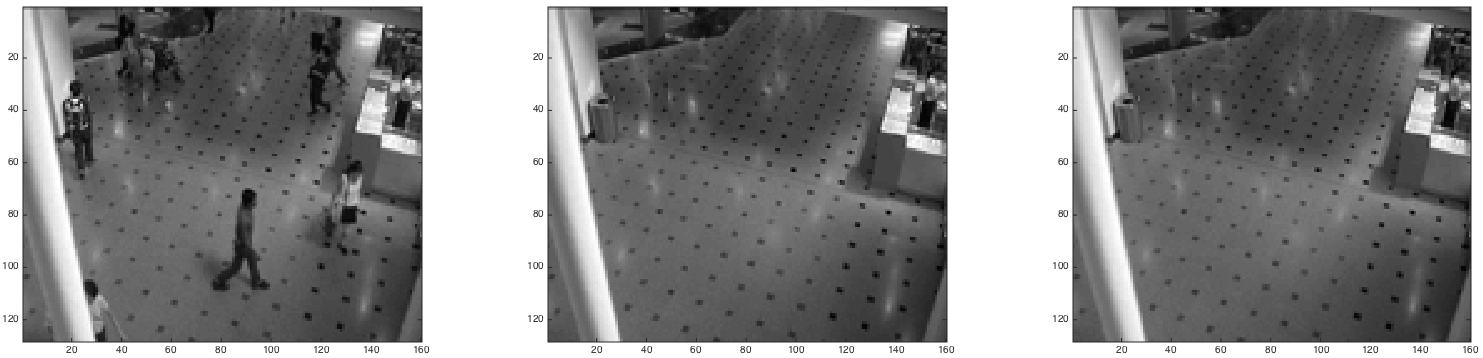}
        \caption{Shopping mall lobby. All the moving people are removed in both the middle and right figures.}
    \end{subfigure}
    \caption{Static background separation from three different videos. a) restaurant food counter, b) airport hallway and c) shopping mall lobby. In each of the cases: the leftmost figure shows one actual frame of the video, the middle figure shows the recovered static background using Robust PCA \cite{candes2011robust} with $\lambda = 1/\sqrt{\max(n,p)}$ and the rightmost figure shows the recovered background using our model with $\lambda = 1/\sqrt{\max(n,p)}$ and $\gamma = 10$. The effect of graph can be appreciated in the first two cases. In a) the changes in the illumination (which are not a part of static background) are more visible in the Robust PCA model than our model. In b) the moving person is more obvious in Robust PCA recovered background than our model. For c) the two models perform equally well.}
    \label{fig:videosfull}
\end{figure*}

\FloatBarrier
\subsection{Parameter Selection for the Proposed Model}

  \begin{figure*}[b]
    \centering
        \centering
        \includegraphics[width=1.0\textwidth]{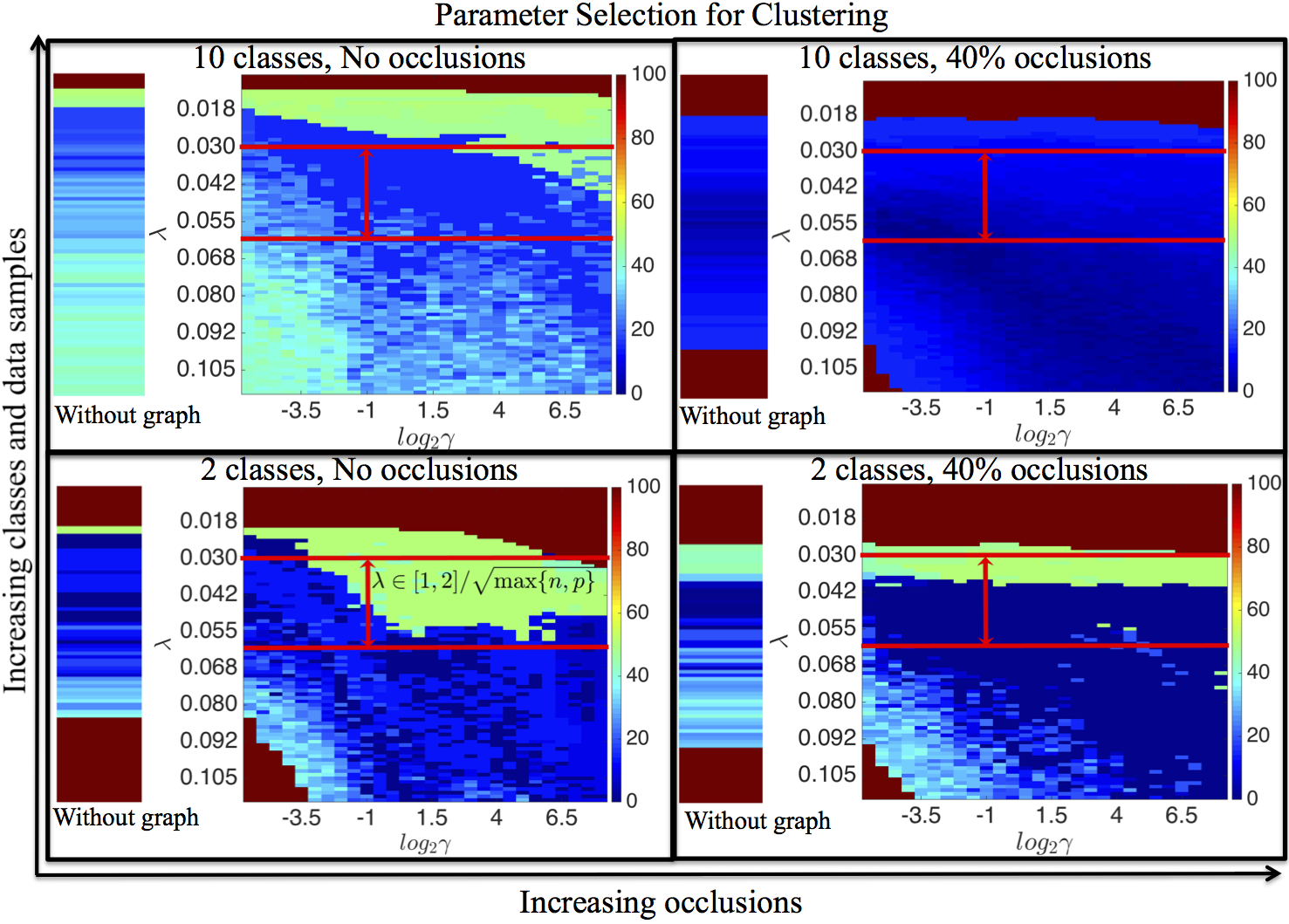}
         \caption{Variation of clustering error over $(\lambda, \gamma)$ grid for different size, classes and occlusions in the CMU PIE data set. x-axis shows the variation with increasing occlusion size and y-axis with increasing size of the data matrix $n$ and number of classes $k$. The horizontal red lines show a range of $\lambda \in [1,2] /\sqrt{\max(n,p)}$. The clustering error always attains a minimum value within this range of $\lambda$ irrespective of the size of the data matrix,  occlusions or the number of classes. Furthermore, a wide range of $\gamma$ values attain a minimum clustering error in the region between the red lines. \textbf{Thus a simple rule can be used to choose a set of good parameters}: \textit{Fix $\lambda \in [1,2] /\sqrt{\max(n,p)}$ and then perform a cross-validation over a coarse  range of $\gamma$ values. In fact $\lambda = 1/\sqrt{\max(n,p)}$ is always a good choice for the images which have no occlusions and are only affected by shadows and illumination changes}. Each strip on the left of $(\lambda, \gamma)$ plot shows the variation of clustering error with $\lambda$ (y-axis) for $\gamma = 0$ (Robust PCA). Clearly, the use of a graph increases the range the possible $\lambda$ values which attain a minimum clustering error.}
        \label{fig:images_ce}
    \end{figure*}
    
\subsubsection{Parameter Selection for Clustering}\label{sec:params_cluster}

\subsubsection{Parameter Selection for Low-Rank Recovery}\label{sec:params_lowrank}   
     \begin{figure*}[!]
    \centering
        \centering
        \includegraphics[width=1.0\textwidth]{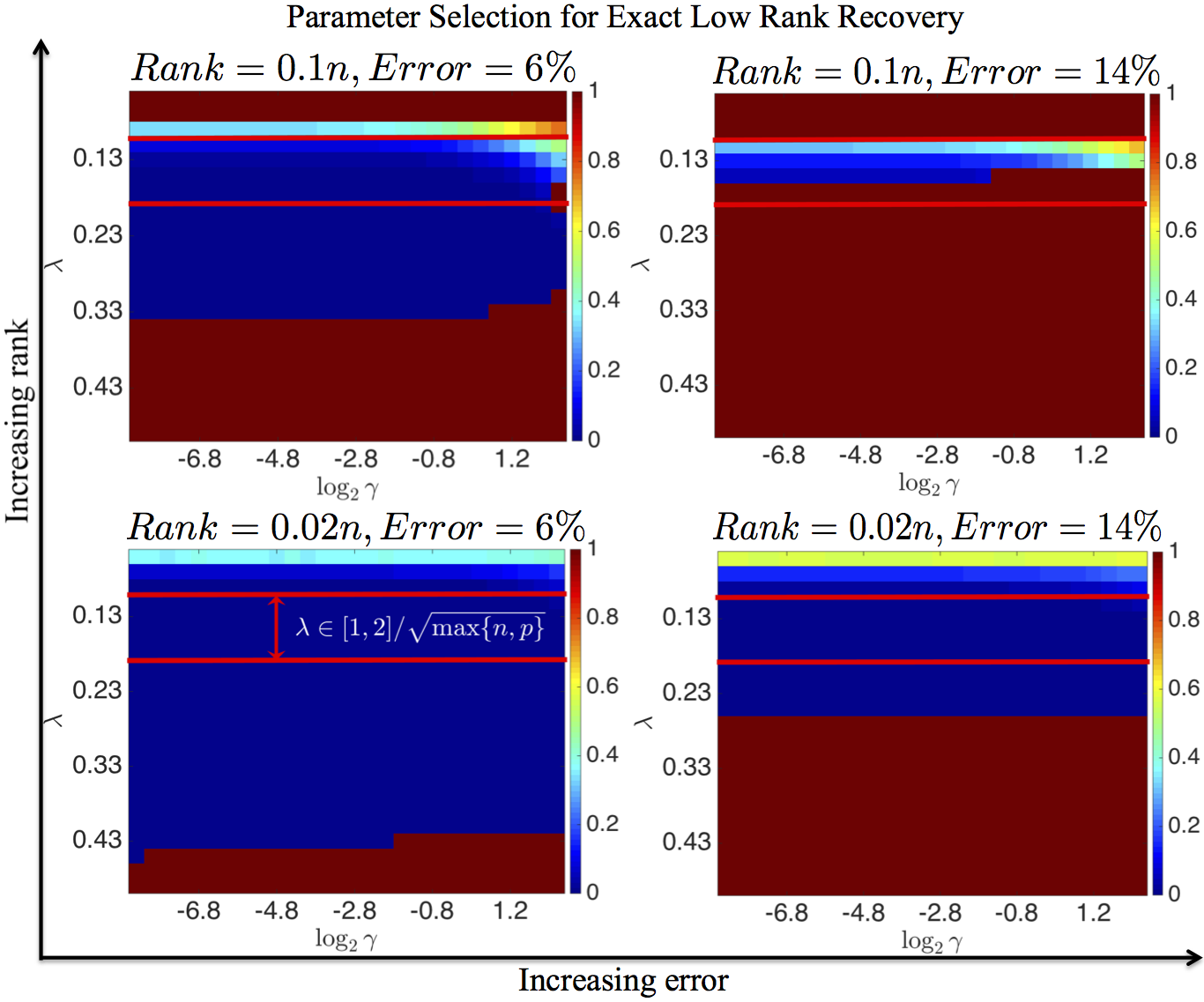}
         \caption{Variation of low-rank normalized reconstruction error over $(\lambda, \gamma)$ grid for different ranks and errors (corruptions) in the artificially generated dataset with Bernoulli support and random sign scheme for sparse errors (Section~\ref{sec:recovery_artificialdata}). x-axis shows the variation with increasing fraction of errors and y-axis with increasing rank of the data matrix $X$. The horizontal red lines show a range of $\lambda \in [1,2] /\sqrt{\max(n,p)}$. The clustering error always attains a minimum value within this range of $\lambda$ irrespective of the size of the $(rank, error)$ setting. Furthermore, a wide range of $\gamma$ values attain a minimum clustering error in the region between the red lines. \textbf{Thus the same simple rule as clustering can be used to choose a set of good parameters}: \textit{Fix $\lambda \in [1,2] /\sqrt{\max(n,p)}$ and then perform a cross-validation over a coarse range of $\gamma$ values}. Similar to our observation for clustering error, $\lambda = 1/\sqrt{\max(n,p)}$ is always a good choice for the datasets with low fraction of errors.}
        \label{fig:images_lowe}
    \end{figure*}

   
   \clearpage
\FloatBarrier
\subsection{Comparison of Computation Times}\label{sec:time}
 \begin{table*}[htbp]
\footnotesize
\caption{A comparison of computational times (in secs) for one run of each model with increasing  size of data matrix. For this experiment, different size of the CMU PIE dataset (n = 300, 600 and 1200) is corrupted with 20\% occlusions and the computation time and accuracy is computed for one run of each model. To perform a fair comparison between the models, the parameter tuple for each model is chosen using the procedure of Section~\ref{sec:param_sel}. The non-convex models are run 10 times and the computational time and accuracy of the run with the minimum clustering error are reported. The convex models are run only once. Clearly,   RGLPCA has the highest computation time, followed by our proposed model. However, the trade-off between the clustering error and computational time is worth observing. Our model takes more  time to converge but attains the minimum clustering error. This large computation time is dominated by the expensive SVD step in every iteration. \textbf{Our future work will be focused on reducing the complexity of this model by exploiting the distributed and parallel computation techniques.} }
\centering
\resizebox{1.0\textwidth}{!}{\begin{tabular}[t]{| c | c | c | c | c | c | c | c |}\hline
 & \textbf{Model} &  \multicolumn{2}{c |}{\textbf{n = 300}}  & \multicolumn{2}{c |}{\textbf{n = 600}}  & \multicolumn{2}{c |}{\textbf{n = 1200}} \\\cline{3-8}
 &                        & \textbf{Time (secs)}   & \textbf{Clustering error (\%)}   & \textbf{Time (secs)}   & \textbf{Clustering error (\%)}   & \textbf{Time (secs)}   & \textbf{Clustering error (\%)}  \\\cline{2-8}
 1  & NCuts &   0.24   &  47.2   &   0.72  &    48.3 &   1.31   &  49.0  \\\hline
 2   & LE     &   0.24   &  48.1    & 0.70   &  50.2    &   1.24    &  52.7  \\\hline
3  &  PCA &  0.11    &  34.3 &   0.17 &  35.9   &  0.20   &   38.3  \\\hline
4  & GLPCA  & 0.12   &  37.5 &   0.43  &   36.1 &    1.60  &  37.2  \\\hline
5  & RGLPCA   &  150.4    & 34.8    &  356.6     &   35.3  &  1187.6    & 37.2  \\\hline
6   &  MMF   &   0.13    &  36.7  &  0.32  & 37.5    &   1.52   & 38.9  \\\hline
7   &  NMF   &  0.15    &  43.9   & 0.62  &   45.6  &   1.30  &  47.4  \\\hline
8   &  GNMF  & 1.20  &   41.2   &  0.62   & 40.3   & 1.30  &  45.9  \\\hline
9   & RPCA    & 59.8   &   10.5   &   159.3   & 12.6  &   678.3  &  14.9 \\\hline
10  &  Our model  &  69.7  &  8.4 &   169.6 &  10.9  &  869.8  &  13.8 \\\hline
\end{tabular}}
\label{tab:times}
\end{table*}

\end{document}